\documentclass[10pt,journal,compsoc]{IEEEtran}
\usepackage{cite}
\ifCLASSINFOpdf

\else

\fi

\usepackage{amsfonts,amssymb}
\usepackage{graphicx}
\usepackage{algorithm}
\usepackage{algorithmic}
\usepackage{amsmath}
\usepackage{booktabs}
\usepackage{multirow}
\usepackage{caption}
\usepackage{subcaption}
\usepackage{color}
\usepackage{ragged2e}
\usepackage{paralist}
\usepackage{enumitem}

\begin{document}

\title{Cross-Domain Visual Matching via Generalized Similarity Measure and Feature Learning}

\author{Liang Lin, Guangrun Wang, Wangmeng Zuo, Xiangchu Feng, and Lei Zhang
\IEEEcompsocitemizethanks {\IEEEcompsocthanksitem
 L. Lin and G. Wang are with School of Data and Computer Science, Sun Yat-sen University, Guangzhou, P. R. China. Email: linliang@ieee.org; wanggrun@mail2.sysu.edu.cn.

\IEEEcompsocthanksitem W. Zuo is with School of Computer Science and Technology, Harbin Institute of Technology, Harbin, P. R. China. Email: cswmzuo@gmail.com.

\IEEEcompsocthanksitem X. Feng is with School of Math. and Statistics, Xidian University, Xi'an, P. R. China. Email: xcfeng@mail.xidian.edu.cn.

\IEEEcompsocthanksitem L. Zhang is with Dept. of Computing, The Hong Kong Polytechnic University, Hong Kong. Email: cslzhang@comp.polyu.edu.hk.}}

\markboth{IEEE TRANSACTIONS ON Pattern Analysis and Machine Intelligence}%
{L. Lin\MakeLowercase{\textit{et al.}}: Cross-Domain Visual Matching via Generalized Similarity Measure and Feature Learning}

\IEEEcompsoctitleabstractindextext{
\begin{abstract}
Cross-domain visual data matching is one of the fundamental problems in many real-world vision tasks, e.g., matching persons across ID photos and surveillance videos. Conventional approaches to this problem usually involves two steps: i) projecting samples from different domains into a common space, and ii) computing (dis-)similarity in this space based on a certain distance. In this paper, we present a novel pairwise similarity measure that advances existing models by i) expanding traditional linear projections into affine transformations and ii) fusing affine Mahalanobis distance and Cosine similarity by a data-driven combination. Moreover, we unify our similarity measure with feature representation learning via deep convolutional neural networks. Specifically, we incorporate the similarity measure matrix into the deep architecture, enabling an end-to-end way of model optimization. We extensively evaluate our generalized similarity model in several challenging cross-domain matching tasks: person re-identification under different views and face verification over different modalities (i.e., faces from still images and videos, older and younger faces, and sketch and photo portraits). The experimental results demonstrate superior performance of our model over other state-of-the-art methods.
\end{abstract}

\begin{IEEEkeywords}
Similarity model, Cross-domain matching, Person verification, Deep learning.
\end{IEEEkeywords}}

\maketitle

\IEEEdisplaynotcompsoctitleabstractindextext

\IEEEpeerreviewmaketitle

\section{Introduction}
\label{sect:intro}

\begin{figure}
\centering
\begin{subfigure}{0.2205\textwidth}
  \centering
  \includegraphics[width=0.93\linewidth]{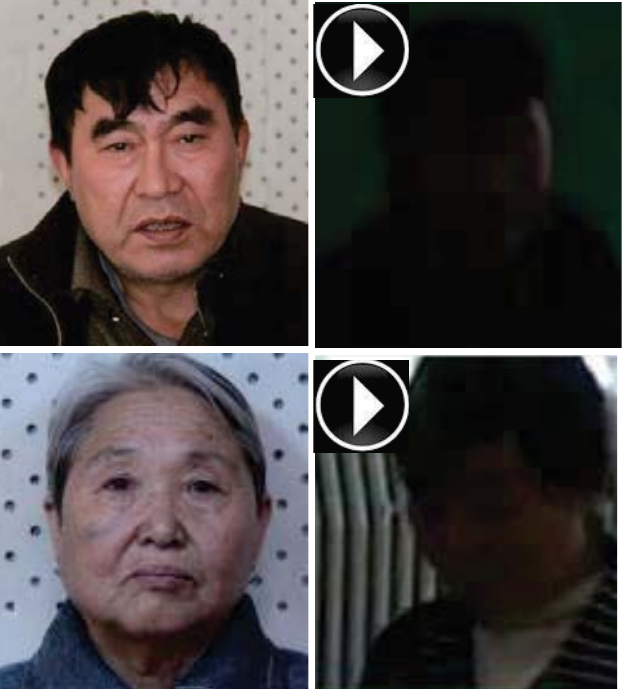}
  \caption{}
\end{subfigure}%
\begin{subfigure}{0.100\textwidth}
  \centering
  \includegraphics[width=0.93\linewidth]{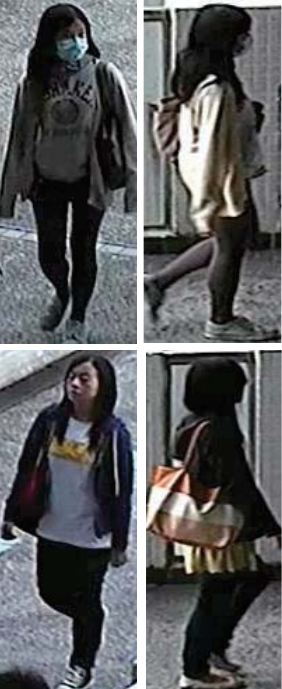}
  \caption{}
\end{subfigure}
\begin{subfigure}{0.158\textwidth}
  \centering
  \includegraphics[width=0.9\linewidth]{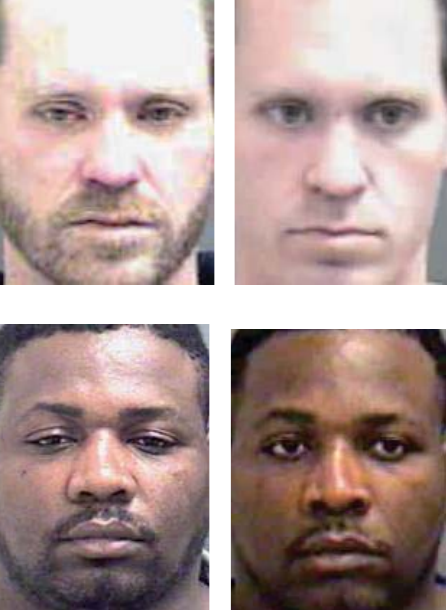}
  \caption{}
\end{subfigure}%
\begin{subfigure}{0.169\textwidth}
  \centering
  \includegraphics[width=0.9\linewidth]{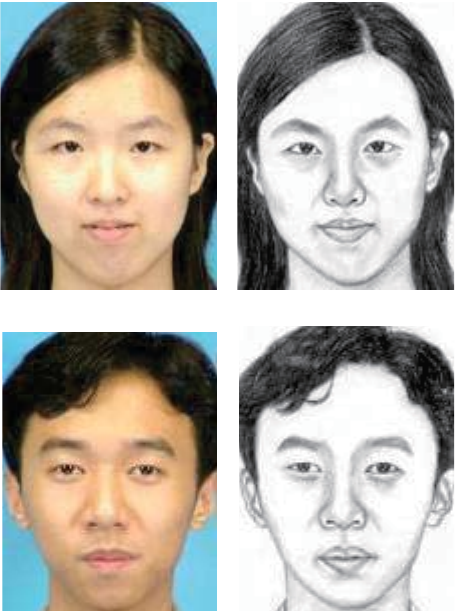}
  \caption{}
\end{subfigure}
\caption{Typical examples of matching cross-domain visual data. (a) Faces from still images and vidoes. (b) Front- and side-view persons. (c) Older and younger faces. (d) Photo and sketch faces.}
\label{fig:datasetexample}
\end{figure}

\IEEEPARstart{V}{isual} similarity matching is arguably considered as one of the most fundamental problems in computer vision and pattern recognition, and this problem becomes more challenging when dealing with cross-domain data. For example, in still-video face retrieval, a newly rising task in visual surveillance, faces from still images captured under a constrained environment are utilized as the queries to find the matches of the same identity in unconstrained videos. Age-invariant and sketch-photo face verification tasks are also examples of cross-domain image matching. Some examples in these applications are shown in Figure \ref{fig:datasetexample}.
%

Conventional approaches (e.g., canonical correlation analysis \cite{hardoon2004canonical} and partial least square regression \cite{sharma2011bypassing}) for cross-domain matching usually follow a procedure of two steps:

\begin{enumerate}[leftmargin=*]
 \item Samples from different modalities are first projected into a common space by learning a transformation. One may simplify the computation by assuming that these cross domain samples share the same projection.
 \item A certain distance is then utilized for measuring the similarity/disimilarity in the projection space. Usually Euclidean distance or inner product are used.
\end{enumerate}

Suppose that $\mathbf{x}$ and $\mathbf{y}$ are two samples of different modalities, and $\mathbf{U}$ and $\mathbf{V}$ are two projection matrices applied on $\mathbf{x}$ and $\mathbf{y}$, respectively. $\mathbf{U}{\mathbf{x}}$ and $\mathbf{V}{\mathbf{y}}$ are usually formulated as linear similarity transformations mainly for the convenience of optimization. A similarity transformation has a good property of preserving the shape of an object that goes through this transformation, but it is limited in capturing complex deformations that usually exist in various real problems, e.g., translation, shearing, and their compositions. On the other hand, Mahalanobis distance, Cosine similarity, and their combination have been widely studied in the research of similarity metric learning, but it remains less investigated on how to unify feature learning and similarity learning, in particular, how to combine Mahalanobis distance with Cosine similarity and integrate the distance metric with deep neural networks for end-to-end learning.

%
%
%

To address the above issues, in this work we present a more general similarity measure and unify it with deep convolutional representation learning. {One of the} key innovations is that we generalize the existing similarity models from two aspects. First, we extend the similarity transformations $\mathbf{U}{\bf{x}}$ and $\mathbf{V}{\bf{y}}$ to the affine transformations by adding a translation vector into them, i.e., replacing $\mathbf{U}{\bf{x}}$ and $\mathbf{V}{\bf{y}}$ with ${{\bf{L}}_{\bf{A}}}{\bf{x}} + {\bf{a}}$ and ${{\bf{L}}_{\bf{B}}}{\bf{y}} + {\bf{b}}$, respectively. Affine transformation is a generalization of similarity transformation without the requirement of preserving the original point in a linear space, and it is able to capture more complex deformations. Second, unlike the traditional approaches choosing either Mahalanobis distance or Cosine similarity, we combine these two measures under the affine transformation. This combination is realized in a data-driven fashion, as discussed in the Appendix, resulting in a novel generalized similarity measure, defined as:

\begin{equation}\label{eq_general_sim}
S(\mathbf{x}, \mathbf{y}) = [\mathbf{x}^T \mbox{ } \mathbf{y}^T \mbox{ } 1] \begin{bmatrix}
 \mathbf{A} \!&\! \mathbf{C}  \!&\! \mathbf{d}\\
\mathbf{C}^T \!&\! \mathbf{B} \!&\! \mathbf{e}\\
\mathbf{d}^T \!&\! \mathbf{e}^T \!&\! f
\end{bmatrix}
\begin{bmatrix} \mathbf{x}\\
\mathbf{y}\\
1
\end{bmatrix},
\end{equation}
where sub-matrices $\mathbf{A}$ and $\mathbf{B}$ are positive semi-definite, representing the self-correlations of the samples in their own domains, and $\mathbf{C}$ is a correlation matrix crossing the two domains.

\begin{figure}
  \centering
  \includegraphics[width=0.40\textwidth]{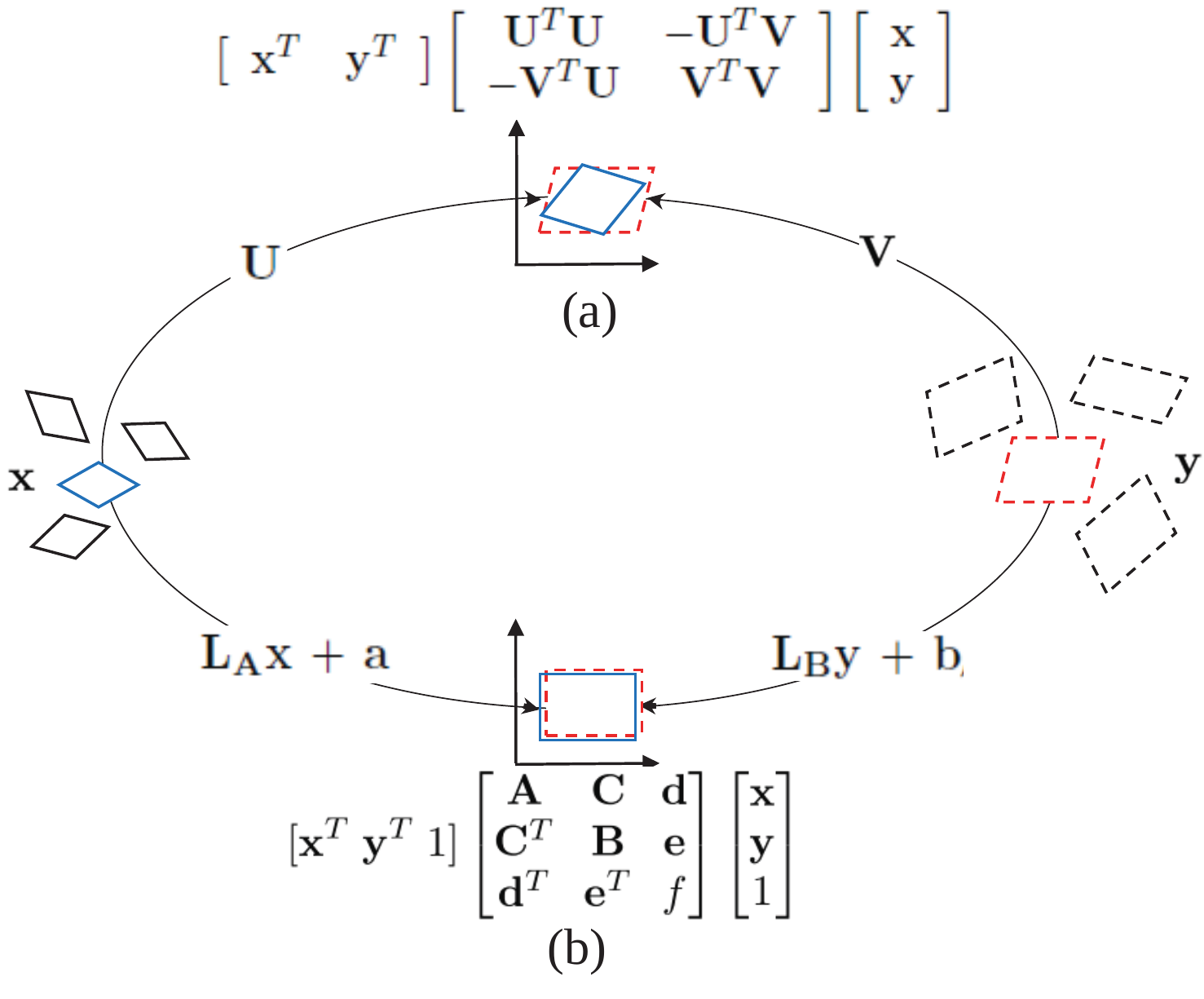}
  \caption{Illustration of the generalized similarity model. Conventional approaches project data by simply using the linear similarity transformations (i.e., $\mathbf{U}$ and $\mathbf{V}$), as illustrated in (a), where Euclidean distance is applied as the distance metric. As illustrated in (b), we improve existing models by i) expanding the traditional linear similarity transformation into an affine transformation and ii) fusing Mahalanobis distance and Cosine similarity. One can see that the case in (a) is a simplified version of our model. Please refer to Appendix section for the deduction details.}
\label{fig:illustration}       
\end{figure}

Figure \ref{fig:illustration} intuitively explains the idea\footnote{Figure \ref{fig:illustration} does not imply that our model geometrically aligns two samples to be matched. Using this example we emphasize the superiority of the affine transformation over the traditional linear similarity transformation on capturing pattern variations in the feature space. }. In this example, it is observed that Euclidean distance under the linear transformation, as (a) illustrates, can be regarded as a special case of our model with $\mathbf{A} = \mathbf{U}^T\mathbf{U}$, $\mathbf{B} = \mathbf{V}^T\mathbf{V}$, $\mathbf{C} = -\mathbf{U}^T\mathbf{V}$, $\mathbf{d} = \mathbf{0}$, $\mathbf{e} = \mathbf{0}$, and $f = 0$. Our similarity model can be viewed as a generalization of several recent metric learning models \cite{li2013learning}\cite{chen2012bayesian}. Experimental results validate that the introduction of $(\mathbf{d}, \mathbf{e}, f)$ and more flexible setting on $(\mathbf{A}, \mathbf{B}, \mathbf{C})$ do improve the matching performance significantly.

Another innovation of this work is that we unify feature representation learning and similarity measure learning. In literature, most of the existing models are performed in the original data space or in a pre-defined feature space, that is, the feature extraction and the similarity measure are studied separately. These methods may have several drawbacks in practice. For example, the similarity models heavily rely on feature engineering and thus lack of generality when handling problems under different scenarios. Moreover, the interaction between the feature representations and similarity measures is ignored or simplified, thus limiting their performances. Meanwhile, deep learning, especially the Convolutional Neural Network (CNN), has demonstrated its effectiveness on learning discriminative features from raw data and benefited to build end-to-end learning frameworks. Motivated by these works, we build a deep architecture to integrate our similarity measure with the CNN-based feature representation learning. Our architecture takes raw images of different modalities as the inputs and automatically produce their representations by sequentially stacking shared sub-network upon domain-specific subnetworks. Upon these layers, we further incorporate the components of our similarity measure by stimulating them with several appended structured neural network layers. The feature learning and the similarity model learning are thus integrated for end-to-end optimization.


In sum, this paper makes three main contributions to cross-domain similarity measure learning.
\begin{itemize}[leftmargin=*]
    \item First, it presents a generic similarity measure by generalizing the traditional linear projection and distance metrics into a unified formulation. Our model can be viewed as a generalization of several existing similarity learning models.
	\item Second, it integrates feature learning and similarity measure learning by building an end-to-end deep architecture of neural networks. Our deep architecture effectively improves the adaptability of learning with data of different modalities.
	\item Third, we extensively evaluate our framework on four challenging tasks of cross-domain visual matching: person re-identification across views\footnote{Person re-identification is arguably a cross-domain matching problem. We introduce it in our experiments since this problem has been receiving increasing attentions recently.}, and face verification under different modalities (i.e., faces from still images and videos, older and younger faces, and sketch and photo portraits). The experimental results {show} that our similarity model outperforms other state-of-the-arts in three of the four tasks and achieves the second best performance in the other one.
\end{itemize}

The rest of the paper is organized as follows. Section \ref{sec:related_work} reviews related work. Section \ref{sec:framework} introduces our generalized similarity model and discusses its connections to existing works. Section \ref{sec:architecture} presents the proposed deep neural network architecture and the learning algorithm in Section \ref{sec:learning_algorithm}. The experimental results, comparisons and ablation studies are presented in Section \ref{sec:experiments}. Section \ref{sec:conclusion} concludes the paper.

\section{Related Work}
\label{sec:related_work}

In literature, to cope with the cross-domain matching of visual data, one can learn a common space for different domains. CCA \cite{hardoon2004canonical} learns the common space via maximizing cross-view correlation, while PLS \cite{sharma2011bypassing} is learned via maximizing cross-view covariance. Coupled information-theoretic encoding is proposed to maximize the mutual information \cite{davis2007information}. Another conventional strategy is to synthesize samples from the input domain into the other domain. Rather than learning the mapping between two domains in the data space, dictionary learning \cite{zhuang2013supervised}\cite{wang2012semi} can be used to alleviate cross-domain heterogeneity, and semi-coupled dictionary learning (SCDL \cite{wang2012semi}) is proposed to model the relationship on the sparse coding vectors from the two domains. Duan et al. proposed another framework called domain adaptation machine (DAM) \cite{duan2012domain} for multiple source domain adaption but they need a set of pre-trained base classifiers.

Various discriminative common space approaches have been developed by utilizing the label information. Supervised information can be employed by the Rayleigh quotient \cite{hardoon2004canonical}, treating the label as the common space \cite{ramage2009labeled}, or employing the max-margin rule \cite{zhu2009medlda}. Using the SCDL framework, structured group sparsity was adopted to utilize the label information \cite{zhuang2013supervised}. Generalization of discriminative common space to multiview was also studied \cite{sharma2012generalized}. Kan et al. proposed a multiview discriminant analysis (MvDA \cite{kan2012multi}) method to obtain a common space for multiple views by optimizing both inter-view and intra-view Rayleigh quotient. In \cite{luo2015learning}, a method to learn shape models using local curve segments with multiple types of distance metrics was proposed.

Moreover, for most existing multiview analysis methods, the target is defined based on the standard inner product or distance between the samples in the feature space. In the field of metric learning, several generalized similarity / distance measures have been studied to improve recognition performance. In \cite{chen2012bayesian}\cite{cao2013similarity}, the generalized distance / similarity measures are formulated as the difference between the distance component and the similarity component to take into account both cross inner product term and two norm terms. Li et al. \cite{li2013learning} adopted the second-order decision function as distance measure without considering the positive semi-definite (PSD) constraint. Chang and Yeung \cite{chang2007ICCV} suggested an approach to learn locally smooth metrics using local affine transformations while preserving the topological structure of the original data. These distance / similarity measures, however, were developed for matching samples from the same domain, and they cannot be directly applied to cross domain data matching.

{To extend traditional single-domain metric learning, Mignon and Jurie \cite{mignon2012ACCV} suggested a cross-modal metric learning (CMML) model, which learns domain-specific transformations based on a generalized logistic loss. Zhai et al. \cite{zhai2013AAAI} incorporated the joint graph regularization with the heterogeneous metric learning model to improve the cross-media retrieval accuracy. In \cite{mignon2012ACCV, zhai2013AAAI}, Euclidean distance is adopted to measure the dissimilarity in the latent space. Instead of explicitly learning domain-specific transformations, Kang et al. \cite{kang2014Arxiv} learned a low rank matrix to parameterize the cross-modal similarity measure by the accelerated proximal gradient (APG) algorithm. However, these methods are mainly based on the common similarity or distance measures and none of them addresses the feature learning problem under the cross-domain scenarios.}

Instead of using hand-crafted features, learning feature representations and contextual relations with deep neural networks, especially the convolutional neural network (CNN) \cite{lecun1989backpropagation}, {has} shown great potential in various pattern recognition tasks such as object recognition \cite{krizhevsky2012imagenet} and semantic segmentation \cite{long2014fully}. { Significant performance gains have also been achieved in face recognition \cite{sun2014deep} and person re-identification \cite{xiong2014person}\cite{zhao2014learning}\cite{ahmed2015improved}\cite{chen2015Arxiv}, mainly attributed to the progress in deep learning.} Recently, several deep CNN-based models have been explored for similarity matching and learning. For example, Andrew et al. \cite{andrew2013deep} proposed a multi-layer CCA model consisting of several stacked nonlinear transformations. Li et al. \cite{li2014deepreid} learned filter pairs via deep networks to handle misalignment, photometric and geometric transforms, and achieved promising results for the person re-identification task. Wang et al. \cite{wang2014learning} learned fine-grained image similarity with deep ranking model. Yi et al. \cite{yi2014deep} presented a deep metric learning approach by generalizing the Siamese CNN. { Ahmed et al. \cite{ahmed2015improved} proposed a deep convolutional architecture to measure the similarity between a pair of pedestrian images. Besides the shared convolutional layers, their network also includes a neighborhood difference layer and a patch summary layer to compute cross-input neighborhood differences. Chen et al. \cite{chen2015Arxiv} proposed a deep ranking framework to learn the joint representation of an image pair and return the similarity score directly, in which the similarity model is replaced by full connection layers.

Our deep model is partially motivated by the above works, and we target on a more powerful solution of cross-domain visual matching by incorporating a generalized similarity function into deep neural networks. Moreover, our network architecture is different from existing works, leading to new state-of-the-art results on several challenging person verification and recognition tasks.}

\section{Generalized Similarity Model}
\label{sec:framework}
In this section, we first introduce the formulation of our deep generalized similarity model and then discuss the connections between our model and existing similarity learning methods.

\subsection{Model Formulation}

According to the discussion in Section \ref{sect:intro}, our generalized similarity measure extends the traditional linear projection and integrates Mahalanobis distance and Cosine similarity into a generic form, as shown in Eqn. (\ref{eq_general_sim}). As we derive in the Appendix, $\mathbf{A}$ and $\mathbf{B}$ in our similarity measure are positive semi-definite but $\mathbf{C}$ does not obey this constraint. Hence, we can further factorize $\mathbf{A}$, $\mathbf{B}$ and $\mathbf{C}$, as:

\begin{equation}\label{eq_factorize}
\begin{matrix}
\mathbf{A} = \mathbf{L_A}^T \mathbf{L_A},\\
\mathbf{B} = \mathbf{L_B}^T \mathbf{L_B},\\
\mathbf{C} = -\mathbf{L_C^x}^T \mathbf{L_C^y}.
\end{matrix}
\end{equation}

Moreover, our model extracts feature representation (i.e., $\mathbf{f}_1(\mathbf{x})$ and $\mathbf{f}_2(\mathbf{y})$) from the raw input data by utilizing the CNNs. Incorporating the feature representation and the above matrix factorization into Eqn. (\ref{eq_general_sim}), we can thus have the following similarity model:

\vspace{-4mm}
\begin{eqnarray}\label{eq_deep_general_sim}
\tilde{S}(\mathbf{x}, \mathbf{y})  \!\!&\!\! = \!\!&\!\! S(\mathbf{f}_1(\mathbf{x}), \mathbf{f}_2(\mathbf{y}))\\\nonumber
 \!\!&\!\! = \!\!&\!\! [\mathbf{f}_1(\mathbf{x})^T \mbox{ } \mathbf{f}_2(\mathbf{y})^T \mbox{ } 1]\!\! \begin{bmatrix}
 \mathbf{A} \!\!&\!\! \mathbf{C} \! \!&\!\! \mathbf{d}\\
\mathbf{C}^T \!\!&\!\! \mathbf{B} \!\!&\!\! \mathbf{e}\\
\mathbf{d}^T \!\!&\!\! \mathbf{e}^T \!\!&\!\! f
\end{bmatrix}\!\!
\begin{bmatrix} \mathbf{f}_1(\mathbf{x})\\
\mathbf{f}_2(\mathbf{y})\\
1
\end{bmatrix}
\\\nonumber
 \!\!&\!\! = \!\!&\!\! \|\mathbf{L_A} \mathbf{f}_1(\mathbf{x})\|^2  +  \|\mathbf{L_B} \mathbf{f}_2(\mathbf{y})\|^2 + 2\mathbf{d}^T \mathbf{f}_1(\mathbf{x}) \\\nonumber
 \!\! &-& 2(\mathbf{L_C^x} \mathbf{f}_1(\mathbf{x}))^T (\mathbf{L_C^y} \mathbf{f}_2(\mathbf{y})) \!+\! 2\mathbf{e}^T \mathbf{f}_2(\mathbf{y}) \!+\! f.
\end{eqnarray}

Specifically, $\mathbf{L_A} \mathbf{f}_1(\mathbf{x})$, $\mathbf{L_C^x} \mathbf{f}_1(\mathbf{x})$, $\mathbf{d}^T \mathbf{f}_1(\mathbf{x})$ can be regarded as the similarity components for $\mathbf{x}$, while $\mathbf{L_B} \mathbf{f}_2(\mathbf{y})$, $\mathbf{L_C^y} \mathbf{f}_2(\mathbf{y})$, $\mathbf{d}^T \mathbf{f}_2(\mathbf{y})$ accordingly for $\mathbf{y}$. These similarity components are modeled as the weights that connect neurons of the last two layers. For example, a portion of output activations represents $\mathbf{L_A} \mathbf{f}_1(\mathbf{x})$ by taking $\mathbf{f}_1(\mathbf{x})$ as the input and multiplying the corresponding weights $\mathbf{L_A}$. In the following, we discuss the formulation of our similarity learning.

The objective of our similarity learning is to seek a function $\tilde{S}({\bf{x}},{\bf{y}})$ that satisfies a set of similarity/disimilarity constraints. Instead of learning similarity function on hand-crafted feature space, we take the raw data as input, and introduce a deep similarity learning framework to integrate nonlinear feature learning and generalized similarity learning. Recall that our deep generalized similarity model is in Eqn. (\ref{eq_general_sim}). $(\mathbf{f}_1(\mathbf{x}), \mathbf{f}_2(\mathbf{y}))$ are the feature representations for samples of different modalities, and we use $\mathbf{W}$ to indicate their parameters. We denote $\mathbf{\Phi} = (\mathbf{L_A}, \mathbf{L_B}, \mathbf{L_C^x}, \mathbf{L_C^y}, \mathbf{d}, \mathbf{e}, f)$ as the similarity components for sample matching. Note that $\tilde{S}(\mathbf{x}, \mathbf{y})$ is asymmetric, i.e., $\tilde{S}({\bf{x}},{\bf{y}}) \ne \tilde{S}({\bf{y}},{\bf{x}})$. This is reasonable for cross-domain matching, because the similarity components are domain-specific.

Assume that $\mathcal{D}=\{(\{\mathbf{x}_{i  }, \mathbf{y}_{i  }\}, \ell_i)\}_{i=1}^N$ is a training set of cross-domain sample pairs, where $\{\mathbf{x}_{i }, \mathbf{y}_{i }\}$ denotes the $i$th pair, and $\ell_i$ denotes the corresponding label of $\{\mathbf{x}_{i }, \mathbf{y}_{i }\}$ indicating whether $\mathbf{x}_{i }$ and $\mathbf{y}_{i }$ are from the same class:

\begin{equation}
\ell_i = \ell(\mathbf{x}_{i }, \mathbf{y}_{i })= \left\{ {\begin{array}{*{20}{c}}
{ - 1,{\kern 1pt} {\kern 1pt} {\kern 1pt} {\kern 1pt} {\kern 1pt} {\kern 1pt} {\kern 1pt} {\kern 1pt} {\kern 1pt} {\kern 1pt} {\kern 1pt} {\kern 1pt} {\kern 1pt} {\kern 1pt} {\kern 1pt} {\kern 1pt} {\kern 1pt} {\kern 1pt} {\kern 1pt} {\kern 1pt} {\kern 1pt} {\kern 1pt} {\kern 1pt} {\kern 1pt} c({\bf{x}}) = c({\bf{y}})}\\
{1,{\kern 1pt} {\kern 1pt} {\kern 1pt} {\kern 1pt} {\kern 1pt} {\kern 1pt} {\kern 1pt} {\kern 1pt} {\kern 1pt} {\kern 1pt} {\kern 1pt} {\kern 1pt} {\kern 1pt} {\kern 1pt} {\kern 1pt} {\kern 1pt} {\kern 1pt} {\kern 1pt} {\kern 1pt} {\kern 1pt} {\kern 1pt} {\kern 1pt} {\kern 1pt} {\kern 1pt} {\kern 1pt} {\kern 1pt} {\kern 1pt} {\kern 1pt} {\kern 1pt} {\kern 1pt} {\kern 1pt} {\kern 1pt} {\kern 1pt} {\kern 1pt} {\kern 1pt} {\kern 1pt} {\kern 1pt} \mbox{otherwise}}
\end{array}} \right.,
\end{equation}
where $c(\mathbf{x})$ denotes the class label of the sample $\mathbf{x}$. An ideal deep similarity model is expected to satisfy the following constraints:

\begin{equation}\label{eq_constraint}
\tilde{S}({\bf{x}}_{i },{\bf{y}}_{i })\left\{ {\begin{array}{*{20}{l}}
{ < -1, \mbox{ } \mbox{if} \mbox{ } \ell_i = -1}\\
{ \ge 1, \mbox{ } \mbox{ } \mbox{ } \mbox{ } \mbox{otherwise}  }
\end{array}} \right.
\end{equation}
for any $\{\mathbf{x}_{i }, \mathbf{y}_{i }\}$.

 Note that the feasible solution that satisfies the above constraints may not exist. To avoid this scenario, we relax the hard constraints in Eqn. (\ref{eq_constraint}) by introducing a hinge-like loss:

\vspace{-4mm}
\begin{equation}\label{eq_hinge}
G(\mathbf{W}, \mathbf{\Phi}) = \sum_{i=1}^{N} (1-\ell_i \tilde{S}(\mathbf{x}_{i }, \mathbf{y}_{i }) )_+.
\end{equation}
To improve the stability of the solution, some regularizers are further introduced, resulting in our deep similarity learning model:

\vspace{-4mm}
\begin{equation}\label{eq_dsl}
({\bf{\hat W}},{\bf{\hat \Phi }}) = \arg {\min _{{\bf{W}},{\bf{\Phi }}}}\sum\limits_{i = 1}^N {(1 - {\ell _i}\tilde S(} {{\bf{x}}_i},{{\bf{y}}_i}){)_ + } + \Psi ({\bf{W}},{\bf{\Phi }}),
\end{equation}
where $\Psi(\mathbf{W}, \mathbf{\Phi}) = \lambda \|\mathbf{W}\|^2 + \mu \|\mathbf{\Phi}\|^2$ denotes the regularizer on the parameters of the feature representation and generalized similarity models.

\subsection{Connection with Existing Models}
\label{sec:discussion}
Our generalized similarity learning model is a generalization of many existing metric learning models, while they can be treated as special cases of our model by imposing some extra constraints on $(\mathbf{A}, \mathbf{B}, \mathbf{C}, \mathbf{d}, \mathbf{e}, f)$.

\begin{figure*}
  \centering
  \includegraphics[width=0.8\textwidth]{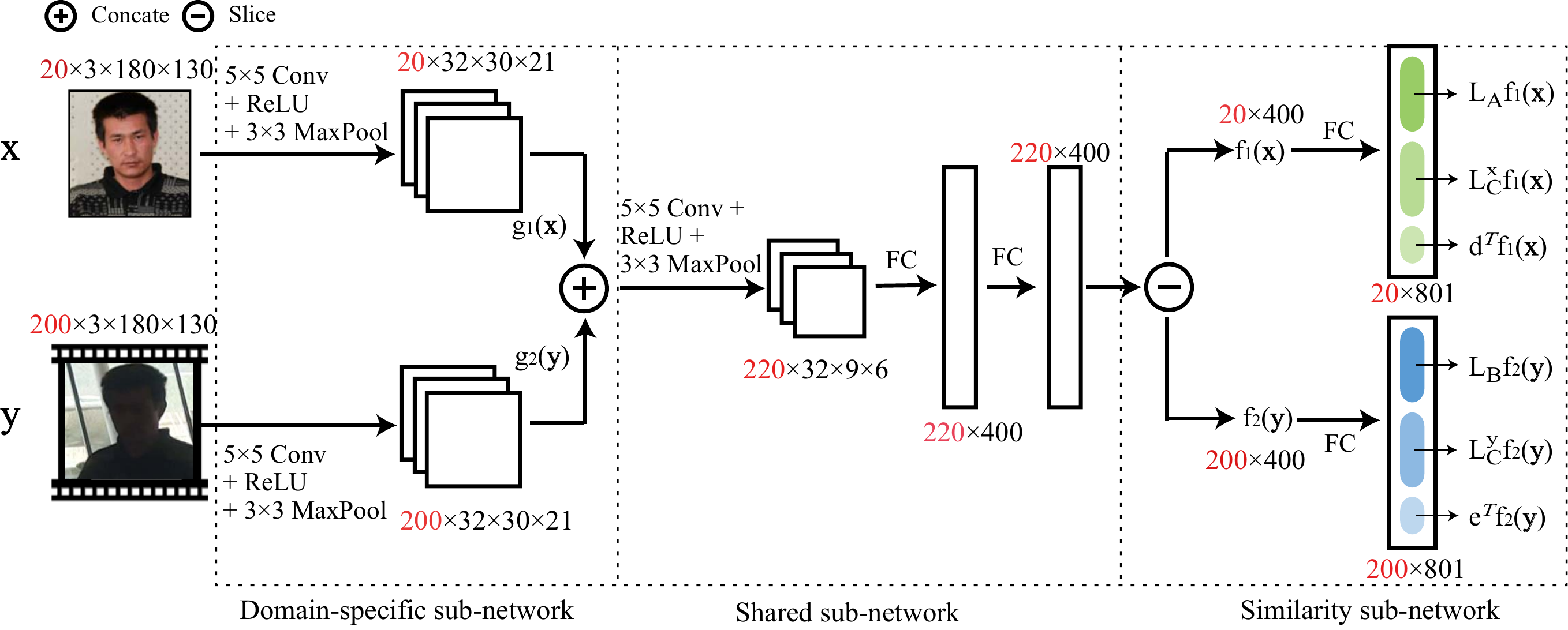}
  \caption{Deep architecture of our similarity model. This architecture is comprised of three parts: domain-specific sub-network, shared sub-network and similarity sub-network. The first two parts extract feature representations from samples of different domains, which are built upon a number of convolutional layers, max-pooling operations and fully-connected layers. The similarity sub-network includes two structured fully-connected layers that incorporate the similarity components in Eqn. (\ref{eq_deep_general_sim}).}
\label{fig:Architecture}       
\end{figure*}

Conventional similarity model usually is defined as $S_{\mathbf{M}}(\mathbf{x}, \mathbf{y}) = \mathbf{x}^T \mathbf{M} \mathbf{y}$, and this form is equivalent to our model, when $\mathbf{A} = \mathbf{B} = 0$, $\mathbf{C} = \frac{1}{2}\mathbf{M}$, $\mathbf{d} = \mathbf{e} = \mathbf{0}$, and $f = 0$. Similarly, the Mahalanobis distance $D_{\mathbf{M}}(\mathbf{x}, \mathbf{y}) = (\mathbf{x} - \mathbf{y})^T \mathbf{M}(\mathbf{x} - \mathbf{y})$ is also regarded as a special case of our model, when $\mathbf{A} = \mathbf{B} = \mathbf{M}$, $\mathbf{C} = \mathbf{-M}$, $\mathbf{d} = \mathbf{e} = \mathbf{0}$, and $f = 0$.

In the following, we connect our similarity model to two state-of-the-art similarity learning methods, i.e., LADF~\cite{li2013learning} and Joint Bayesian~\cite{chen2012bayesian}.

In \cite{li2013learning}, Li et al. proposed to learn a decision function that jointly models a distance metric and a locally adaptive thresholding rule, and the so-called LADF (i.e., Locally-Adaptive Decision Function) is formulated as a second-order large-margin regularization problem. Specifically, LADF is defined as:

\vspace{-3mm}
\begin{equation}\label{eq_LADF}
F({\bf{x}},{\bf{y}}) \!=\! {{\bf{x}}^T}{\bf{Ax}} \!+\! {{\bf{y}}^T}{\bf{Ay}} \!+\! 2{{\bf{x}}^T}{\bf{Cy}} \!+\! {{\bf{d}}^T}{\bf{(x \!+\! y)}} \!+\! f.
\end{equation}
One can observe that $F({\bf{x}},{\bf{y}}) = S({\bf{x}},{\bf{y}})$ when we set $\mathbf{B} = \mathbf{A}$ and $\mathbf{e} = \mathbf{d}$ in our model.

It should be noted that LADF treats $\mathbf{x}$ and $\mathbf{y}$ using the same metrics, i.e., $\mathbf{A}$ for both $\bf{x}^T \bf{Ax}$ and ${{\bf{y}}^T}{\bf{Ay}}$, and $\mathbf{d}$ for $\mathbf{d}^T\mathbf{x}$ and $\mathbf{d}^T\mathbf{y}$. Such a model is reasonable for matching samples with the same modality, but may be unsuitable for cross-domain matching where $\bf{x}$  and $\bf{y}$ are with different modalities. Compared with LADF, our model uses $\mathbf{A}$ and $\mathbf{d}$ to calculate $\bf{x}^T\bf{Ax}$ and $\bf{d}^T\bf{x}$, and uses $\mathbf{B}$ and $\mathbf{e}$ to calculate ${{\bf{y}}^T}{\bf{By}}$ and $\bf{e}^T\bf{y}$, making our model more effective for cross-domain matching.

In \cite{chen2012bayesian}, Chen et al. extended the classical Bayesian face model by learning a joint distributions (i.e., intra-person and extra-person variations) of sample pairs. Their decision function is posed as the following form:

\begin{equation}\label{eq_BFR}
J(\mathbf{x}, \mathbf{y}) = {{\bf{x}}^T}{\bf{Ax + }}{{\bf{y}}^T}{\bf{Ay - 2}}{{\bf{x}}^T}{\bf{Gy}}.
\end{equation}
Note that the similarity metric model proposed in \cite{cao2013similarity} also adopted such a form. Interestingly, this decision function is also a special variant of our model by setting $\mathbf{B} = \mathbf{A}$, $\mathbf{C} = -\mathbf{G}$, $\mathbf{d} = \mathbf{0}$, $\mathbf{e} = \mathbf{0}$, and $\mathbf{f} = 0$.

In summary, our similarity model can be regarded as the generalization of many existing cross-domain matching and metric learning models, and it is more flexible and suitable for cross-domain visual data matching.

\section{Joint Similarity and Feature Learning}
\label{sec:architecture}
In this section, we introduce our deep architecture that integrates the generalized similarity measure with convolutional feature representation learning.

\subsection{Deep Architecture}

As discussed above, our model defined in Eqn. (\ref{eq_dsl}) jointly handles similarity function learning and feature learning. This integration is achieved by building a deep architecture of convolutional neural networks, which is illustrated in Figure \ref{fig:Architecture}. It is worth mentioning that our architecture is able to handle the input samples of different modalities with unequal numbers, e.g., $20$ samples of $\mathbf{x}$ and $200$ samples of $\mathbf{y}$ are fed into the network in a way of batch processing.

From left to right in Figure \ref{fig:Architecture}, two domain-specific sub-networks $\mathbf{g}_1(\mathbf{x})$ and $\mathbf{g}_2(\mathbf{y})$ are applied to the samples of two different modalities, respectively. Then, the outputs of $\mathbf{g}_1(\mathbf{x})$ and $\mathbf{g}_2(\mathbf{y})$ are concatenated into a shared sub-network $\mathbf{f}(\cdot)$. We make a superposition of $\mathbf{g}_1(\mathbf{x})$ and $\mathbf{g}_2(\mathbf{y})$ to feed $\mathbf{f}(\cdot)$. At the output of $\mathbf{f}(\cdot)$, the feature representations of the two samples are extracted separately as  $\mathbf{f}_1(\mathbf{x})$ and $\mathbf{f}_2(\mathbf{y})$, which is indicated by the slice operator in Figure \ref{fig:Architecture}. Finally, these learned feature representations are utilized in the structured fully-connected layers that incorporate the similarity components defined in Eqn. (\ref{eq_deep_general_sim}). In the following, we introduce the detailed setting of the three sub-networks.

{\bf Domain-specific sub-network.}  We separate two branches of neural networks to handle the samples from different domains. Each network branch includes one convolutional layer with $3$ filters of size $5 \times 5$ and the stride step of $2$ pixels. The rectified nonlinear activation is utilized. Then, we follow by a one max-pooling operation with size of $3 \times 3$ and its stride step is set as $3$ pixels.

{\bf Shared sub-network.} For this component, we stack one convolutional layer and two fully-connected layers. The convolutional layer contains $32$ filters of size $5\times 5$ and the filter stride step is set as $1$ pixel. The kernel size of the max-pooling operation is $3\times 3$ and its stride step is $3$ pixels. The output vectors of the two fully-connected layers are of $400$ dimensions. We further normalize the output of the second fully-connected layer before it is fed to the next sub-network.

{\bf Similarity sub-network.} A slice operator is first applied in this sub-network, which partitions the vectors into two groups corresponding to the two domains. For the example in Figure \ref{fig:Architecture}, $220$ vectors are grouped into two sets, i.e., $\mathbf{f}_1(\mathbf{x})$ and $\mathbf{f}_2(\mathbf{y})$, with size of $20$ and $200$, respectively. $\mathbf{f}_1(\mathbf{x})$ and $\mathbf{f}_2(\mathbf{y})$ are both of $400$ dimensions. Then, $\mathbf{f}_1(\mathbf{x})$ and $\mathbf{f}_2(\mathbf{y})$ are fed to two branches of neural network, and each branch includes a fully-connected layer. We divide the activations of these two layers into six parts according to the six similarity components. As is shown in Figure \ref{fig:Architecture}, in the top branch the neural layer connects to $\mathbf{f}_1(\mathbf{x})$ and outputs { ${{\bf{L}}_{\bf{A}}}{\bf{f_1(x)}}$, ${{\bf{L}}^{\bf{x}}_{\bf{C}}}{\bf{f_1(x)}}$, and ${{\bf{d}}^T}{\bf{f_1(x)}}$, respectively.} {In the bottom branch, the layer outputs ${{\bf{L}}_{\bf{B}}}{\bf{f_2(y)}}$, ${{\bf{L}}^{\bf{y}}_{\bf{C}}}{\bf{f_2(y)}}$, and ${{\bf{e}}^T}{\bf{f_2(y)}}$, respectively, by connecting to $\mathbf{f}_2(\mathbf{y})$. In this way, the similarity measure is tightly integrated with the feature representations, and they can be jointly optimized during the model training.} { Note that ${f}$ is a parameter of the generalized similarity measure in Eqn. (\ref{eq_general_sim}). Experiments show that the value of ${f}$ only affects the learning convergence rather than the matching performance. {Thus we empirically set ${f}= -1.9$ in our experiments}.}

In the deep architecture, we can observe that the similarity components of $\mathbf{x}$ and those of $\mathbf{y}$ do not interact to each other by the factorization until the final aggregation calculation, that is, computing the components of $\mathbf{x}$ is independent of $\mathbf{y}$. This leads to a good property of efficient matching. In particular, for each sample stored in a database, we can pre-computed its feature representation and the corresponding similarity components, and the similarity matching in the testing stage will be very fast.

%

%
\subsection{Model Training}
\label{sec:learning_algorithm}

In this section, we discuss the learning method for our similarity model training. To avoid loading all images into memory, we use the mini-batch learning approach, that is, in each training iteration, a subset of the image pairs are fed into the neural network for model optimization.

For notation simplicity in discussing the learning algorithm, we start by introducing the following definitions:{

\vspace{-4mm}
\begin{equation}\label{eq_forshort}
\begin{array}{rl}
{\tilde{{\bf{x}}}} \buildrel \Delta \over = & {{\bf{[}}\begin{array}{*{20}{c}}
{{{\bf{L}}_{\bf{A}}}{\bf{f_1(x)}}}&{{{\bf{L}}^{\bf{x}}_{\bf{C}}}{\bf{f_1(x)}}}&{{{\bf{d}}^T}{\bf{f_1(x)}}}
\end{array}{\bf{]}}^T}\\

{\tilde{{\bf{y}}}} \buildrel \Delta \over = & {{\bf{[}}\begin{array}{*{20}{c}}
{{{\bf{L}}_{\bf{B}}}{\bf{f_2(y)}}}&{{{\bf{L}}^{\bf{y}}_{\bf{C}}}{\bf{f_2(y)}}}&{{{\bf{e}}^T}{\bf{f_2(y)}}}
\end{array}{\bf{]}}^T},
\end{array}
\end{equation}
}where $\tilde{\bf{x}}$ and $\tilde{\bf{y}}$ denote the output layer's activations of the samples ${\bf x}$ and ${\bf y}$. Prior to incorporating Eqn. (\ref{eq_forshort}) into the similarity model in Eqn. (\ref{eq_deep_general_sim}), we introduce three transformation matrices (using Matlab representation):

\begin{equation}\label{eq_pdef}
\begin{array}{*{20}{l}}
{{{\bf{P}}_1} = \left[ {\begin{array}{*{20}{c}}
{{{\bf{I}}^{r \times r}}}&{{{\bf{0}}^{r \times (r + 1)}}}
\end{array}} \right],}\\
{{{\bf{P}}_2} = \left[ {\begin{array}{*{20}{c}}
{{{\bf{0}}^{r \times r}}}&{{{\bf{I}}^{r \times r}}}&{{{\bf{0}}^{r \times 1}}}
\end{array}} \right],}\\
{{{\bf{p}}_3} = \left[ {\begin{array}{*{20}{c}}
{{{\bf{0}}^{1 \times 2r}}}&{{1^{1 \times 1}}}
\end{array}} \right]^T,}
\end{array}
\end{equation}
where ${r}$ equals to the dimension of the output of shared neural network (i.e., the dimension of ${f(\bf x)}$ and ${f(\bf y)}$), an ${\bf{I}}$ indicates the identity matrix. Then, our similarity model can be re-written as:
\begin{equation}\label{eq_simila}
\begin{array}{*{20}{c}}
{\tilde S({\bf{x}},{\bf{y}}) = {{({{\bf{P}}_1} {\tilde{\bf{x}}})}^T}{{\bf{P}}_1} \tilde{\bf{x}} + {{({{\bf{P}}_1} \tilde{\bf{y}})}^T}{{\bf{P}}_1} \tilde{\bf{y}} - 2{{({{\bf{P}}_2} \tilde{\bf{x}})}^T}{{\bf{P}}_2} {\tilde{\bf{y}}}}\\
{ + 2{{\bf{p}}_3^T} {\tilde{\bf{x}}} + 2{{\bf{p}}_3^T} {\tilde{\bf{y}}} + f}.
\end{array}
\end{equation}

{
Incorporating Eqn. (\ref{eq_simila}) into the loss function Eqn. (\ref{eq_hinge}), we have the following objective:

\vspace{-4mm}
\begin{equation}\label{eq_derivebatch}
\begin{aligned}
\begin{array}{l}
G({\bf{W}},{\bf{\Phi }}; \mathcal{D})\\
 = \sum\limits_{i = 1}^N {} \left\{ {} \right.1 - {\ell _i}\left[ {} \right.{({{\bf{P}}_1}{\widetilde {\bf{x}}_i})^T}{{\bf{P}}_1}{\widetilde {\bf{x}}_i} + {({{\bf{P}}_1}{\widetilde {\bf{y}}_i})^T}{{\bf{P}}_1}{\widetilde {\bf{y}}_i} - \\
2{({{\bf{P}}_2}{\widetilde {\bf{x}}_i})^T}{{\bf{P}}_2}{\widetilde {\bf{y}}_i} + 2{{\bf{p}}_3^T}{\widetilde {\bf{x}}_i} + 2{{\bf{p}}_3^T}{\widetilde {\bf{y}}_i} + f\left. {} \right]{\left. {} \right\}_ + }
\end{array}
\end{aligned},
\end{equation}
where the summation term denotes the hinge-like loss for the cross domain sample pair $\{ {\tilde{\bf{x}}_i},{\tilde{\bf{y}}_i}\} $, ${N}$ is the total number of pairs, ${\bf{W}}$ represents the feature representation of different domains and ${\bf{\Phi}}$ represents the similarity model. ${\bf{W}}$ and ${\bf{\Phi}}$ are both embedded as weights connecting neurons of layers in our deep neural network model, as Figure \ref{fig:Architecture} illustrates. 

The objective function in Eqn. (\ref{eq_derivebatch}) is defined in sample-pair-based form. To optimize it using SGD, one should apply a certain scheme to generate mini-batches of the sample pairs, which usually costs much computation and memory. Note that the sample pairs in training set $\mathcal{D}$ are constructed from the original set of samples from different modalities $\mathcal{Z} = \{ \{\mathcal{X}\}, \{\mathcal{Y}\} \}$, where $\mathcal{X} = \{ {{\bf{x}}^1}, ..., {{\bf{x}}^j}, ..., {{\bf{x}}^{M_{\bf{x}}}} \}$ and $\mathcal{Y} = \{ {{\bf{y}}^1}, ..., {{\bf{y}}^j}, ..., {{\bf{y}}^{M_{\bf{y}}}} \}$. The superscript denotes the sample index in the original training set, e.g., ${{\bf{x}}^j} \in \mathcal{X} = \{ {{\bf{x}}^1}, ..., {{\bf{x}}^j}, ..., {{\bf{x}}^{M_{\bf{x}}}} \}$ and ${{\bf{y}}^j} \in \mathcal{Y} = \{ {{\bf{y}}^1}, ..., {{\bf{y}}^j}, ..., {{\bf{y}}^{M_{\bf{y}}}} \}$, while the subscript denotes the index of sample pairs, e.g., ${{\bf{x}}_i} \in \{ {{\bf{x}}_i},{{\bf{y}}_i}\}  \in {\cal D}$. ${{M_{\bf{x}}}}$ and ${{M_{\bf{y}}}}$ denote the total number of samples from different domains. Without loss of generality, we define ${{\bf{z}}^j} = {{\bf{x}}^j}$ and ${{\bf{z}}^{M_{\bf{x}}+j}} = {{\bf{y}}^j}$. For each pair $\{ {{\bf{x}}_i},{{\bf{y}}_i}\}$ in $\mathcal{D}$, we have ${\bf{z}}^{j_{i,1}} = {\bf{x}}_i$ and ${\bf{z}}^{j_{i,2}} = {\bf{y}}_i $ with $1 \leq {j_{i,1}} \leq M_{\bf{x}}$ and $M_{\bf{x}}+1 \leq {j_{i,2}} \leq M_{\bf{z}} (= M_{\bf{x}}+M_{\bf{y}})$. And we also have ${\widetilde {\bf{z}}}^{j_{i,1}} = {\widetilde {\bf{x}}}_i$ and ${\widetilde {\bf{z}}}^{j_{i,2}} = {\widetilde {\bf{y}}}_i$.

Therefore, we rewrite Eqn. (\ref{eq_derivebatch}) in a sample-based form:

\vspace{-4mm}
\begin{equation}\label{eq_batch}
\begin{array}{l}
L({\bf{W}},{\bf{\Phi }}; \mathcal{Z}) \\
 = \sum\limits_{i = 1}^N {} \left\{ {} \right.1 - {\ell _i}\left[ {} \right.{({{\bf{P}}_1}{\widetilde {\bf{z}}^{j_{i,1}}})^T}{{\bf{P}}_1}{\widetilde {\bf{z}}^{j_{i,1}}} + {({{\bf{P}}_1}{\widetilde {\bf{z}}^{j_{i,2}}})^T}{{\bf{P}}_1}{\widetilde {\bf{z}}^{j_{i,2}}} - \\
2{({{\bf{P}}_2}{\widetilde {\bf{z}}^{j_{i,1}}})^T}{{\bf{P}}_2}{\widetilde {\bf{z}}^{j_{i,2}}} + 2{{\bf{p}}_3^T}{\widetilde {\bf{z}}^{j_{i,1}}} + 2{{\bf{p}}_3^T}{\widetilde {\bf{z}}^{j_{i,2}}} + f\left. {} \right]{\left. {} \right\}_ + }
\end{array},
\end{equation}
Given ${\bf{\Omega }} = ({\bf{W}},{\bf{\Phi }})$, the loss function in Eqn. (\ref{eq_dsl}) can also be rewritten in the sample-based form:

\vspace{-4mm}
\begin{equation}\label{eq_simplified}
{{H}}({\bf{\Omega }}) = {{L}}({\bf{\Omega }};{\mathcal{Z}}) + \Psi ({\bf{\Omega }}).
\end{equation}
{The objective in Eqn. (\ref{eq_simplified}) can be optimized by the mini-batch back propagation algorithm}. Specifically, we update the parameters by gradient descent:

\vspace{-4mm}
\begin{equation}\label{eq_update}
{\bf{\Omega }} = {\bf{\Omega }} - \alpha \frac{\partial }{{\partial {\bf{\Omega }}}}{{H}}({\bf{\Omega }}),
\end{equation}
where $\alpha $ denotes the learning rate. The key problem of solving the above equation is calculating $\frac{\partial }{{\partial\bf{\Omega }}}{{L(}}{\bf{\Omega }}{\rm{)}}$. As is discussed in \cite{ding2015deep}, there are two ways to this end, i.e., pair-based gradient descent and sample-based gradient descent. Here we adopt the latter to reduce the requirements on computation and memory cost.

Suppose a mini-batch of training samples $\{{\bf{z}}^{j_{1,{\bf{x}}}}, ..., {\bf{z}}^{j_{n_{\bf{x}},{\bf{x}}}}, {\bf{z}}^{j_{1,{\bf{y}}}}, ..., {\bf{z}}^{j_{n_{\bf{y}},{\bf{y}}}} \}$ from the original set ${\mathcal{Z}}$, where $1 \leq {j_{i,{\bf{x}}}} \leq M_{\bf{x}}$ and $M_{\bf{x}}+1 \leq {j_{i,{\bf{y}}}} \leq M_{\bf{z}}$. Following the chain rule, calculating the gradient for all pairs of samples is equivalent to summing up the gradient for each sample, 

\vspace{-4mm}
\begin{equation}\label{eq_samplebasedgrad}
\frac{\partial }{{{{\partial\bf{\Omega }}}}}{{L}}({\bf{\Omega }}) = \sum\limits_{j} {\frac{{\partial {{L}}}}{{\partial{\tilde{\bf{z}}^{j}}}} \frac{{\partial {\tilde{\bf{z}}^{j}}}}{{{{\partial\bf{\Omega }}}}}} ,
\end{equation}
where ${j}$ can be either ${j_{i,{\bf{x}}}}$ or ${j_{i,{\bf{y}}}}$.

Using ${\bf{z}}^{j_{i,{\bf{x}}}}$ as an example, we first introduce an indicator function ${\bf{1}}_{{\bf{z}}^{j_{i,{\bf{x}}}}}({\bf{z}}^{j_{i,{\bf{y}}}})$ before calculating the partial derivative of output layer activation for each sample $\frac{{\partial {{L}}}} {{\partial{\tilde{\bf{z}}^{j_{i,{\bf{x}}}}}}}$. Specifically, we define ${\bf{1}}_{{\bf{z}}^{j_{i,{\bf{x}}}}}({\bf{z}}^{j_{i,{\bf{y}}}}) = 1$ when $\{ {{\bf{z}}^{j_{i,{\bf{x}}}}}, {\bf{z}}^{j_{i,{\bf{y}}}} \}$ is a sample pair and $\ell_{{j_{i,{\bf{x}}}}, {j_{i,{\bf{y}}}}} \tilde S({{\bf{z}}^{j_{i,{\bf{x}}}}}, {\bf{z}}^{j_{i,{\bf{y}}}}) < 1$. Otherwise we let ${\bf{1}}_{{\bf{z}}^{j_{i,{\bf{x}}}}}({\bf{z}}^{j_{i,{\bf{y}}}}) = 0$. $\ell_{{j_{i,{\bf{x}}}}, {j_{i,{\bf{y}}}}}$, indicating where ${\bf{z}}^{j_{i,{\bf{x}}}}$ and ${\bf{z}}^{j_{i,{\bf{y}}}}$ are from the same class. With ${\bf{1}}_{{\bf{z}}^{j_{i,{\bf{x}}}}}({\bf{z}}^{j_{i,{\bf{y}}}})$, the gradient of ${\tilde {{\bf{z}}^{j_{i,{\bf{x}}}}}}$ can be written as

\vspace{-3mm}
\begin{equation}\label{eq_gradimgz}
\frac{\partial L}{{\partial {\tilde{\bf{z}}^{j_{i,{\bf{x}}}}}}} \!\!=\!\!  - \sum\limits_{{j_{i,{\bf{y}}}}} {2 {\bf{1}}_{{\bf{z}}^{j_{i,{\bf{x}}}}}({\bf{z}}^{j_{i,{\bf{y}}}}) {\ell_{{j_{i,{\bf{x}}}}, {j_{i,{\bf{y}}}}}} ({{\bf{P}}_1^T}{{\bf{P}}_1}{\tilde{\bf{z}}^{j_{i,{\bf{x}}}}} \!-\! {{\bf{P}}_2^T}{{\bf{P}}_2}{\tilde{\bf{z}}^{j_{i,{\bf{y}}}}} \!+\! {{\bf{p}}_3})}.
\end{equation}
The calculation of $\frac{\partial L}{{\partial {\tilde{\bf{z}}^{j_{i,{\bf{y}}}}}}}$ can be conducted in a similar way. The algorithm of calculating the partial derivative of output layer activation for each sample is shown in Algorithm \ref{alg:gradimg}.
}

\renewcommand{\algorithmicrequire}{ \textbf{Input:}} 
\renewcommand{\algorithmicensure}{ \textbf{Output:}} 
\renewcommand{\algorithmicforall}{\textbf{for each}}
\begin{small}
\begin{algorithm}[htb]
\caption{Calculate the derivative of the output layer's activation for each sample}
\label{alg:gradimg}
\begin{algorithmic}[1]
\REQUIRE ~~\\
   The output layer's activation for all samples
\ENSURE ~~\\
The partial derivatives of output layer's activation for all the samples
\FORALL {sample ${{\bf{z}}^j}$}
 	\STATE Initialize the partner set ${{\mathcal{M}}^j}$ containing the sample ${{\bf{z}}^j}$ with ${{\mathcal{M}}^j} = \emptyset $;
	\FORALL {pair $\{ {{\bf{x}}_i},{{\bf{y}}_i}\} $}
		\IF {pair $\{ {{\bf{x}}_i},{{\bf{y}}_i}\} $ contains  the sample ${{\bf{z}}^j}$}
            \IF {pair $\{ {{\bf{x}}_i},{{\bf{y}}_i}\} $ satisfies ${\ell _i} \tilde S({\bf{x}}_i,{\bf{y}}_i) < 1$}
                \STATE  ${{\mathcal{M}}^i} \leftarrow $ $\{ {{\mathcal{M}}^i}$, the corresponding partner of ${{\bf{z}}^j}$ in $ \{ {{\bf{x}}_i},{{\bf{y}}_i}\} \} $;
		    \ENDIF
		\ENDIF
	\ENDFOR
	\STATE Compute the derivatives for the sample ${{\bf{z}}^j}$ with all the partners in ${{\mathcal{M}}^j}$,  and sum these derivatives to be the desired partial derivative for sample ${{\bf{z}}^j}$'s output layer's activation using Eqn. (\ref{eq_gradimgz});
\ENDFOR

\end{algorithmic}
\end{algorithm}
\end{small}

{
Note that all the three sub-networks in our deep architecture are differentiable. We can easily use the back-propagation procedure~\cite{lecun1989backpropagation} to compute the partial derivatives with respect to the hidden layers and model parameters ${\bf{\Omega }}$. We summarize the overall procedure of deep generalized similarity measure learning in Algorithm \ref{alg:batchTraining}.

If all the possible pairs are used in training, the sample-based form allows us to generate $n_{\bf{x}} \times n_{\bf{y}}$ sample pairs from a mini-batch of $n_{\bf{x}} + n_{\bf{y}}$. On the other hand, the sample-pair-based form may require $2 n_{\bf{x}} n_{\bf{y}}$ samples or less to generate $n_{\bf{x}} \times n_{\bf{y}}$ sample pairs. In gradient computation, from Eqn. (\ref{eq_gradimgz}), for each sample we only require calculating ${{\bf{P}}_1^T}{{\bf{P}}_1}{\tilde{\bf{z}}^{j_{i,{\bf{x}}}}}$ once and ${{\bf{P}}_2^T}{{\bf{P}}_2}{\tilde{\bf{z}}^{j_{i,{\bf{y}}}}}$ $n_{\bf{y}}$ times in the sample-based form. While in the sample-pair-based form, ${{\bf{P}}_1^T}{{\bf{P}}_1}{\tilde{\bf{z}}^{j_{i,{\bf{x}}}}}$ and ${{\bf{P}}_2^T}{{\bf{P}}_2}{\tilde{\bf{z}}^{j_{i,{\bf{y}}}}}$ should be computed $n_{\bf{x}}$ and $n_{\bf{y}}$ times, respectively. In sum, the sample-based form generally results in less computation and memory cost.
}

\renewcommand{\algorithmicrequire}{ \textbf{Input:}} 
\renewcommand{\algorithmicensure}{ \textbf{Output:}} 
\begin{small}
\begin{algorithm}[htb]
\caption{Generalized Similarity Learning}
\label{alg:batchTraining}
\begin{algorithmic}[1]
\REQUIRE ~~\\
   Training set, initialized parameters $\mathbf{W}$ and $ \mathbf{\Phi}$, learning rate {$\alpha$}, $t \leftarrow 0$ \\
\ENSURE ~~\\
  Network parameters $\mathbf{W}$ and $ \mathbf{\Phi}$
\WHILE {$t <= T$}
  \STATE Sample training pairs $\mathcal{D}$;
  \STATE Feed the sampled images into the network;
  \STATE Perform a feed-forward pass for all the samples and compute the net activations for each sample ${{\bf{z}}^i}$;
  \STATE {Compute the partial derivative of the output layer's activation for each sample by Algorithm \ref{alg:gradimg}.}
  \STATE {Compute the partial derivatives of the hidden layers' activations for each sample following the chain rule;}
  \STATE Compute the desired gradients $\frac{\partial }{{\partial {\bf{\Omega }}}}{{H}}({\bf{\Omega }})$ using the back-propagation procedure;
  \STATE Update the parameters using Eqn. (\ref{eq_update});
\ENDWHILE

\end{algorithmic}
\end{algorithm}
\end{small}

{\bf Batch Process Implementation.}~Suppose that the training image set is divided into $K$ categories, each of which contains $O_1$ images from the first domain and $O_2$ images from the second domain. Thus we can obtain a maximum number $(K \times {O_1}) \times (K \times {O_2})$ of pairwise samples, which is quadratically more than the number of source images $K \times (O_1 +O_2)$. In real application, since the number of stored images may reach millions, it is impossible to load all the data for network training. To overcome this problem, we implement our learning algorithm in a batch-process manner. Specifically, in each iteration, only a small subset of cross domain image pairs are generated and fed to the network for training. According to our massive experiments, randomly generating image pairs is infeasible, which may cause the image distribution over the special batch becoming scattered, making valid training samples for a certain category very few and degenerating the model. Besides, images in any pair are almost impossible to come from the same class, making the positive samples very few. In order to overcome this problem, an effective cross domain image pair generation scheme is adopted to train our generalized similarity model. For each round, we first randomly choose $\widehat{K}$ instance categories. For each category,  a number of $\widehat{O_1}$ images first domain and a number of $\widehat{O_2}$ from second domain are randomly selected. For each selected images in first domain, we randomly take samples from the second domain and the proportions of positive and negative samples are equal. In this way, images distributed over the generated samples are relatively centralized and the model will effectively converge.


%
\section{Experiments}
\label{sec:experiments}

In this section, we apply our similarity model in four representative tasks of matching cross-domain visual data and adopt several benchmark datasets for evaluation: i) person re-identification under different views on CUHK03 \cite{li2014deepreid} and CUHK01 \cite{li2012human} datasets; ii) age-invariant face recognition on MORPH \cite{ricanek2006morph}, { CACD \cite{chen2014cross} and CACD-VS \cite{chen2015CACD-VS} datasets}; iii) sketch-to-photo face matching on CUFS dataset \cite{wang2009face}; iv) face verification over still-video domains on COX face dataset \cite{CoxFace}. On all these tasks, state-of-the-art methods are employed to compare with our model.


{{\em Experimental setting.}} Mini-batch learning is adopted in our experiments to save memory cost. {In each task, we randomly select a batch of sample from the original training set to generate a number of pairs (e.g., $4800$).} The initial parameters of the convolutional and the full connection layers are set by two zero-mean {Gaussian Distributions}, whose standard deviations are $0.01$ and $0.001$ respectively. Other specific settings to different tasks are included in the following sub-sections.

{ In addition, {ablation studies are presented to reveal the benefit of each main component of our method}, e.g., the generalized similarity measure and the joint optimization of CNN feature representation and metric model. We also implement several variants of our method by simplifying the similarity measures for comparison.}


\subsection{Person Re-identification}

Person re-identification, aiming at matching {pedestrian images} across multiple non-overlapped cameras, has attracted increasing attentions in surveillance. Despite that considerable efforts have been made, it is still an open problem due to
the dramatic variations caused by viewpoint and pose changes. To evaluate this task, CUHK03 \cite{li2014deepreid} dataset and CUHK01 \cite{li2012human} dataset are adopted {in our experiments}.

CUHK03 dataset \cite{li2014deepreid} is one of the largest databases for person re-identification. It contains 14,096 images of 1,467 pedestrians collected from 5 different pairs of camera views. Each person is observed by two disjoint camera views and has an average of 4.8 images in each view. We follow the standard setting of using CUHK03 to randomly partition this dataset for 10 times, and a training set (including 1,367 persons) and a testing set (including 100 persons) are obtained without overlap.

{CUHK01 dataset \cite{li2012human} contains 971 individuals, each having two samples from disjoint cameras. Following the setting in \cite{li2014deepreid}\cite{ahmed2015improved}, we partition this dataset into a training set and a testing set: 100 individuals for testing and the others for training.}

{For evaluation on these two benchmarks}, the testing set is further randomly divided into a gallery set of 100 images (i.e., one image per person) and a probe set (including images of individuals from different camera views in contrast to the gallery set) without overlap for 10 times. {We use Cumulative Matching Characteristic (CMC) \cite{gray2007evaluating} as the evaluation metric in this task.}

In our model training, all of the images are resized to $250 \times 100$, and cropped to the size of $230 \times 80$ at the center with a small random perturbation. During every round of learning, 4800 pairs of samples are constructed by selecting 60 persons (or classes) and constructing 80 pairs for each person (class). {For CUHK01, due to each individual only have two samples, the 80 pairs per individual will contain some duplicated pairs.}
\begin{figure}
\centering
\begin{subfigure}{0.33\textwidth}
  \centering
  \includegraphics[width=1\linewidth]{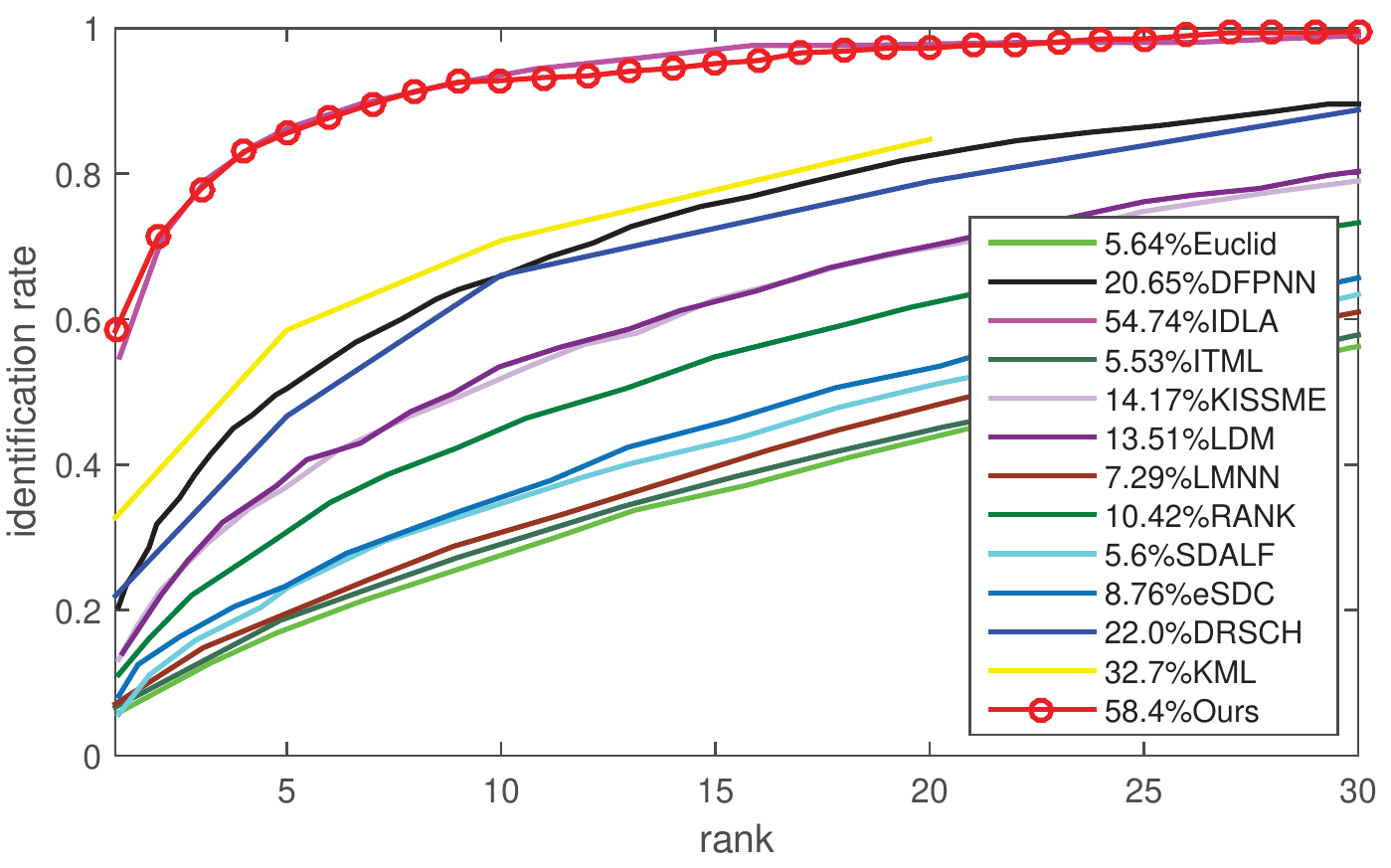}
  \caption{CUHK03}
  \label{fig:sub1}
\end{subfigure}%
\hfill
\begin{subfigure}{0.33\textwidth}
  \centering
  \includegraphics[width=1\linewidth]{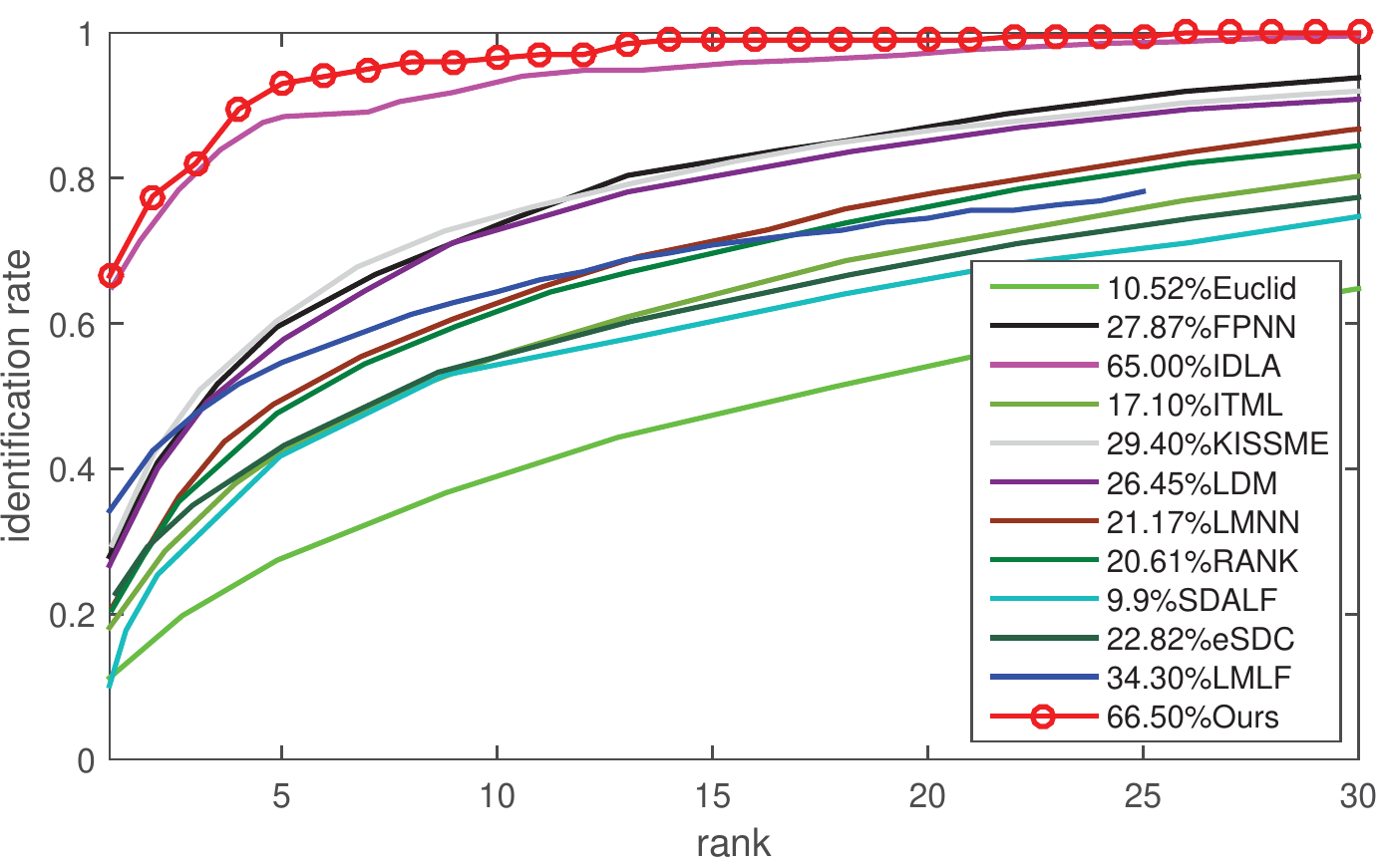}
  \caption{CUHK01}
  \label{fig:sub2}
\end{subfigure}
\caption{CMC curves on (a) CUHK03 \cite{li2014deepreid} dataset and (b) CUHK01 \cite{li2012human} for evaluating person re-identification. Our method has superior performances over existing state-of-the-arts overall.}
\label{fig:reid}
\end{figure}


{\bf Results on CUHK03.} We compare our approach with several state-of-the-art methods, which can be grouped into three categories. First, we adopt five distance metric learning methods based on fixed feature representation, i.e. the Information Theoretic Metric Learning (ITML) \cite{davis2007information}, the Local Distance Metric Learning (LDM) \cite{guillaumin2009you}, the Large Margin Nearest Neighbors (LMNN) \cite{weinberger2005distance}, the learning-to-rank method (RANK) \cite{mcfee2010metric}, and the Kernel-based Metric Learning method (KML) \cite{xiong2014person}. Following their implementation, the handcrafted features of dense color histograms and dense SIFT uniformly sampled from patches are adopted. Second, three methods specially designed for person re-identification are employed in the experiments: SDALF \cite{farenzena2010person}, KISSME \cite{kostinger2012large}, and eSDC \cite{zhao2013unsupervised}. Moreover, several recently proposed deep learning methods, including DRSCH \cite{zhang2015bit}, DFPNN \cite{li2014deepreid} and IDLA \cite{ahmed2015improved}, are also compared with our approach. DRSCH \cite{zhang2015bit} is a supervised hashing framework for integrating CNN feature and hash code learning, while DFPNN and IDLA have been introduced in Section \ref{sec:related_work}.

{
The results are reported in Fig. \ref{fig:reid} (a). {It is encouraging to see that our approach significantly outperforms the competing methods (e.g., improving the state-of-the-art rank-1 accuracy from 54.74\% (IDLA \cite{ahmed2015improved}) to 58.39\%)}. Among the competing methods, ITML \cite{davis2007information}, LDM \cite{guillaumin2009you}, LMNN \cite{weinberger2005distance}, RANK \cite{mcfee2010metric}, KML \cite{xiong2014person}, SDALF \cite{farenzena2010person}, KISSME \cite{kostinger2012large}, and eSDC \cite{zhao2013unsupervised} are all based on hand-crafted features. And the superiority of our approach against them should be attributed to the deployment of both deep CNN features and generalized similarity model. DRSCH \cite{zhang2015bit}, DFPNN \cite{li2014deepreid} and IDLA \cite{ahmed2015improved} adopted CNN for feature representation, but their matching metrics are defined based on traditional linear transformations.
}

{{\bf Results on CUHK01.} Fig. \ref{fig:reid} (b) shows the results of our method and the other competing approaches on CUHK01. In addition to those used on CUHK03, one more method, i.e. LMLF \cite{zhao2014learning}, is used in the comparison experiment. LMLF \cite{zhao2014learning} learns mid-level filters from automatically discovered patch clusters. According to the quantitative results, our method achieves a new state-of-the-art with a rank-1 accuracy of 66.50\%.}


\subsection{Age-invariant Face Recognition}

{Age invariant face recognition is to decide whether two images with different ages belong to the same identity.} The key challenge is to handle the large intra-subject variations caused by aging process while distinguishing different identities. {Other factors, such as illumination, pose, and expression, make age invariant face recognition more difficult.} We conduct the experiments using three datasets, i.e., MORPH \cite{ricanek2006morph}, CACD \cite{chen2014cross}, and CACD-VS \cite{chen2015CACD-VS}.

MORPH \cite{ricanek2006morph} contains more than 55,000 face images of 13,000 individuals, whose ages range from 16 to 77. The average number of images per individual is 4. The training set consists of 20,000 face images from 10,000 subjects, with each subject having two images with the largest age gap. The test set is composed of a gallery set and a probe set from the remaining 3,000 subjects. The gallery set is composed of the youngest face images of each subject. The probe set is composed of the oldest face images of each subject. This experimental setting is the same with those adopted in \cite{gong2013hidden} and \cite{chen2014cross}.

CACD \cite{chen2014cross} is a large scale dataset released in 2014, which contains more than 160,000 images of 2,000 celebrities. We adopt a subset of 580 individuals from the whole database in our experiment, in which we manually remove the noisy images. Among these 580 individuals, the labels of images from 200 individuals have been originally provided, and we annotate the rest of the data. CACD includes large variations not only in pose, illumination, expression but also in ages. {Based on CACD, a verification subset called CACD-VS \cite{chen2015CACD-VS} is further developed, which contains 2,000 positive pairs and 2,000 negative pairs. The setting and testing protocol of CACD-VS are similar to the well-known LFW benchmark \cite{huang2007labeled}, except that CACD-VS contains much more samples for each person.}

All of the images are resized to $200 \times 150$. For data augmentation, images are cropped to the size of $180 \times 130$ at the center with a small random perturbation when feeding to the neural network. Sample-based mini-batch setting is adopted, and 4,800 pairs are constructed for each iteration.

{\bf Results on MORPH.} We compare our method with several state-of-the-art methods, including topological dynamic Bayesian network (TDBN) \cite{bouchaffra2012mapping}, cross-age reference coding {(CARC) \cite{chen2014cross}}, probabilistic hidden factor analysis (HFA) \cite{gong2013hidden}, multi-feature discriminant analysis (MFDA) \cite{li2011discriminative} and 3D aging model \cite{park2010age}. The results are reported in Table \ref{tab:8}(a). {Thanks to the use of CNN representation and generalized similarity measure, our method achieves the recognition rate of 94.35\%, and significantly outperforms the competing methods}.

\begin{figure}
  \centering
  \includegraphics[width=0.33\textwidth]{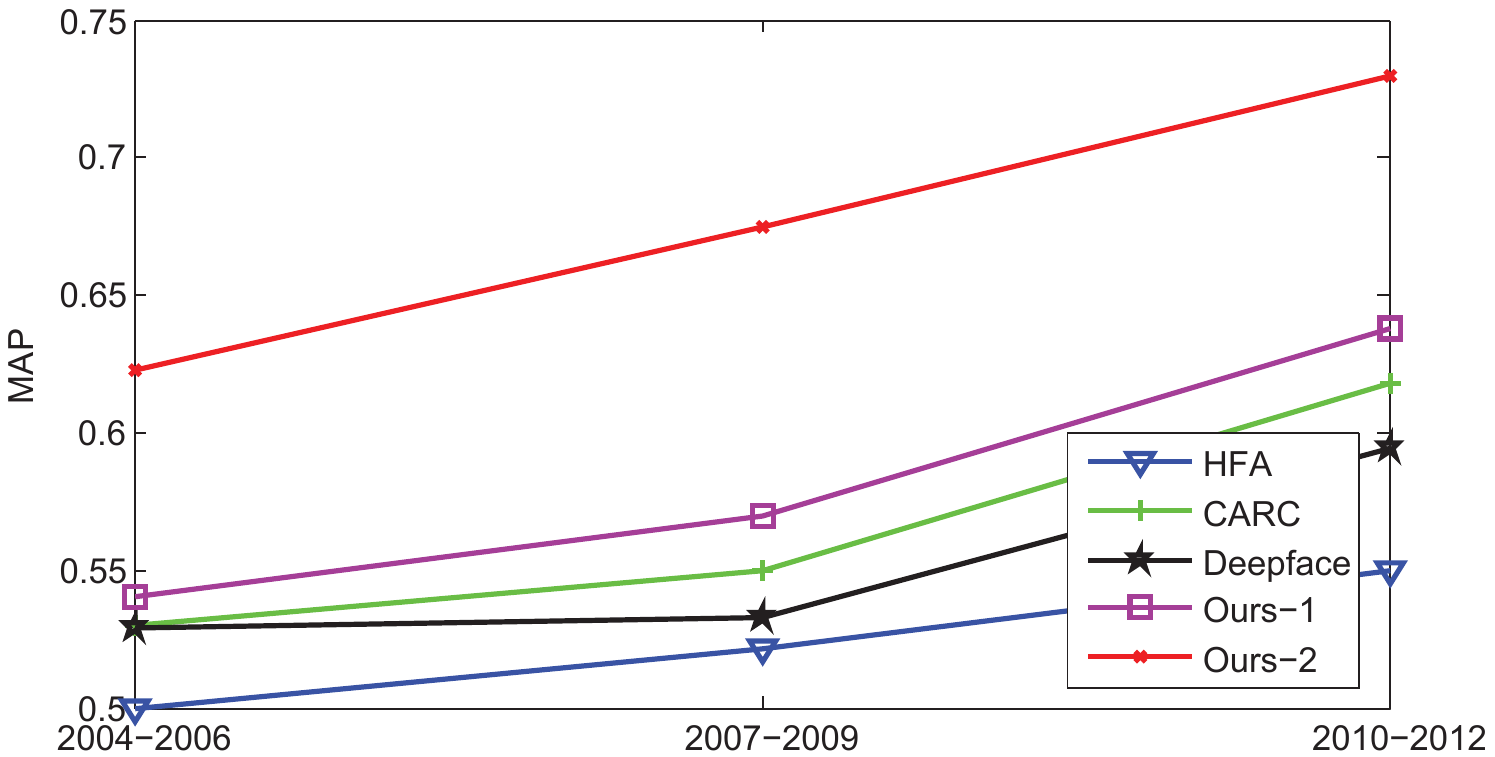}
  \caption{{The retrieval performances on CACD dataset for age-invariant face recognition. Ours-1 and Ours-2 are our method, while the latter uses more training samples.}}
\label{fig:fic_cacd}       
\end{figure}

\begin{table}
\caption{ {Experimental results for age-invariant face recognition.}}\label{tab:8}
\begin{center}
\begin{tabular}{|l|c|}
\multicolumn{2}{c}{(a) Recognition rates on the MORPH dataset.}\\
\hline
Method & Recognition rate\\
\hline\hline
TDBN \cite{bouchaffra2012mapping} & 60\%\\
3D Aging Model \cite{park2010age} & 79.8\%\\
MFDA \cite{li2011discriminative}  & 83.9\%\\
HFA \cite{gong2013hidden}  & 91.1\% \\
CARC \cite{chen2014cross}  & 92.8\% \\
Ours & \textbf{94.4\%} \\
\hline
\end{tabular}
\end{center}
\begin{center}
\begin{tabular}{|l|c|}

\multicolumn{2}{c}{{(b) Verification accuracy on the CACD-VS dataset.}}\\
\hline
Method & verification accuracy\\
\hline\hline
HD-LBP \cite{chen2013blessing} & 81.6\%\\
HFA \cite{gong2013hidden} & 84.4\%\\
CARC \cite{chen2014cross}  & 87.6\% \\
Deepface \cite{taigman2014deepface}  & 85.4\% \\
Ours & \textbf{89.8\%} \\
\hline
\end{tabular}
\end{center}
\end{table}

{\bf Results on CACD.} On this dataset, the protocol is to retrieve face images of the same individual from gallery sets by using a probe set, where the age gap between probe face images and gallery face images is large. Following the experimental setting in \cite{chen2014cross}, we set up 4 gallery sets according to the years when the photos were taken: $[2004-2006]$, $[2007-2009]$, $[2010-2012]$, and $[2013]$. And we use the set of $[2013]$ as the probe set to search for matches in the rest of three sets. {{We introduce several state-of-the-art methods for comparison, including CARC \cite{chen2014cross}, HFA \cite{gong2013hidden} and one deep learning based method, Deepface \cite{taigman2014deepface}}. The results of CARC \cite{chen2014cross} and HFA \cite{gong2013hidden} are borrowed from their papers. The results of Deepface \cite{taigman2014deepface} and our approach (i.e., Ours-1) are implemented based on the 200 originally annotated individuals, where 160 samples are used for model training. From the quantitative results reported in Figure \ref{fig:fic_cacd}, our model achieves superior performances over the competing methods. Furthermore, we also report the result of our method (i.e., Ours-2) by using images of 500 individuals as training samples. One can see that, the performance of our model can be further improved by increasing training data.}

{{\bf Results on CACD-VS.} {Following the setting in \cite{chen2015CACD-VS}, we further evaluate our approach by conducting the general face verification experiment.} Specifically, for all of the competing methods, we train the models on CACD and test on CACD-VS, and the optimal threshold value for matching is obtained by exhaustive search. The results produced by our methods and the others (i.e., CARC \cite{chen2014cross}, HFA \cite{gong2013hidden}, HD-LBP \cite{chen2013blessing} and Deepface \cite{taigman2014deepface}) are reported in  Table \ref{tab:8} (b). It is worth mentioning that our method improves the state-of-the-art recognition rate from 87.6\% (by CARC \cite{chen2014cross} \cite{taigman2014deepface}) to 89.8\%. {Thanks to the introduction of generalized similarity measure our approach achieves higher verification accuracy than Deepface. Note that an explicit face alignment was adopted in \cite{taigman2014deepface} before the CNN feature extraction, which is not in our framework. }


\subsection{Sketch-photo Face Verification}

Sketch-photo face verification is an interesting yet challenging task, which aims to verify whether a face photo and a drawing face sketch belong to the same individual. This task has an important application of assisting law enforcement, i.e., using face sketch to find candidate face photos. It is however difficult to match photos and sketches in two different modalities. For example, hand-drawing may bring unpredictable face distortion and variation compared to the real photo, and face sketches often lack of details that can be important cues for preserving identity.

We evaluate our model on this task using the CUFS dataset \cite{wang2009face}. There are 188 face photos in this dataset, in which 88 are selected for training and 100 for testing. Each face has a corresponding sketch that is drawn by the artist. All of these face photos are taken at frontal view with a normal lighting condition and neutral expression.

\begin{figure*}
\centering
\begin{subfigure}{0.25\textwidth}
  \centering
  \includegraphics[width=1\linewidth]{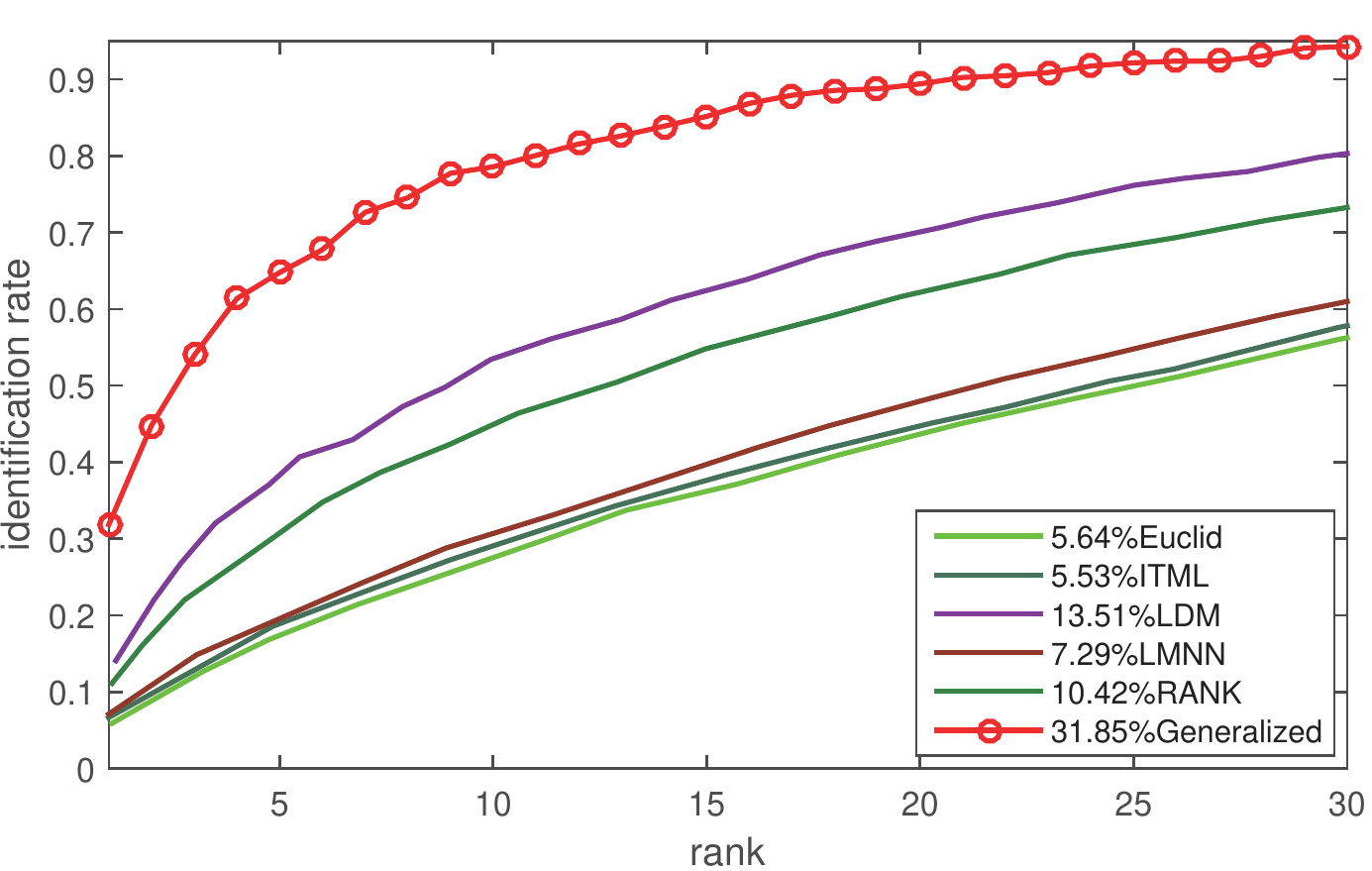}
  \caption{sim.+hand.fea,\\
    CUHK03}
  \label{fig:a1}
\end{subfigure}%
\begin{subfigure}{0.25\textwidth}
  \centering
  \includegraphics[width=1\linewidth]{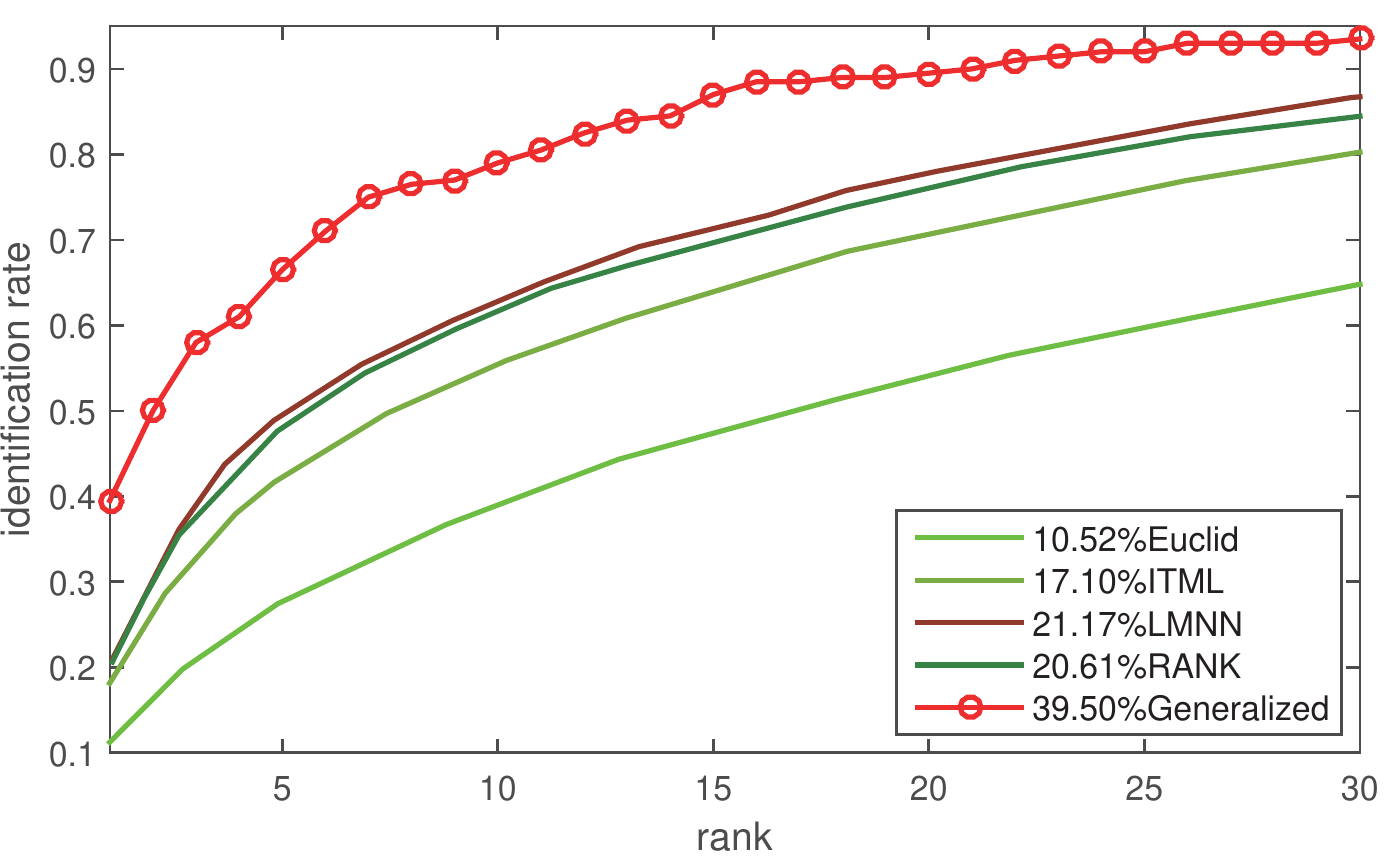}
  \caption{sim.+hand.fea,\\
    CUHK01}
  \label{fig:a2}
\end{subfigure}
\begin{subfigure}{0.26\textwidth}
  \centering
  \includegraphics[width=1\linewidth]{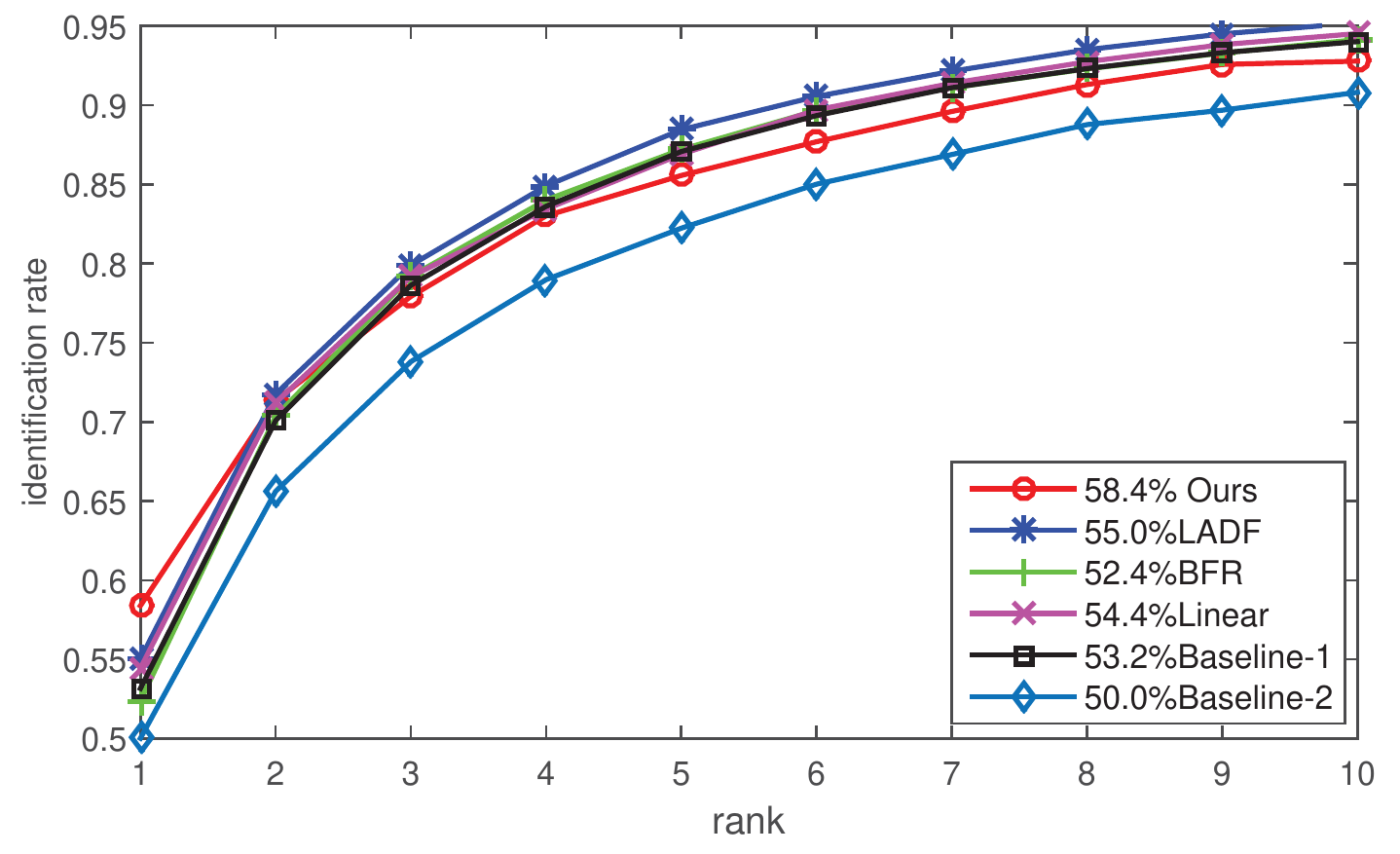}
  \caption{sim.+deep.fea\\
   CUHK03}
  \label{fig:a3}
\end{subfigure}%
\hfill
\begin{subfigure}{0.26\textwidth}
  \centering
  \includegraphics[width=1\linewidth]{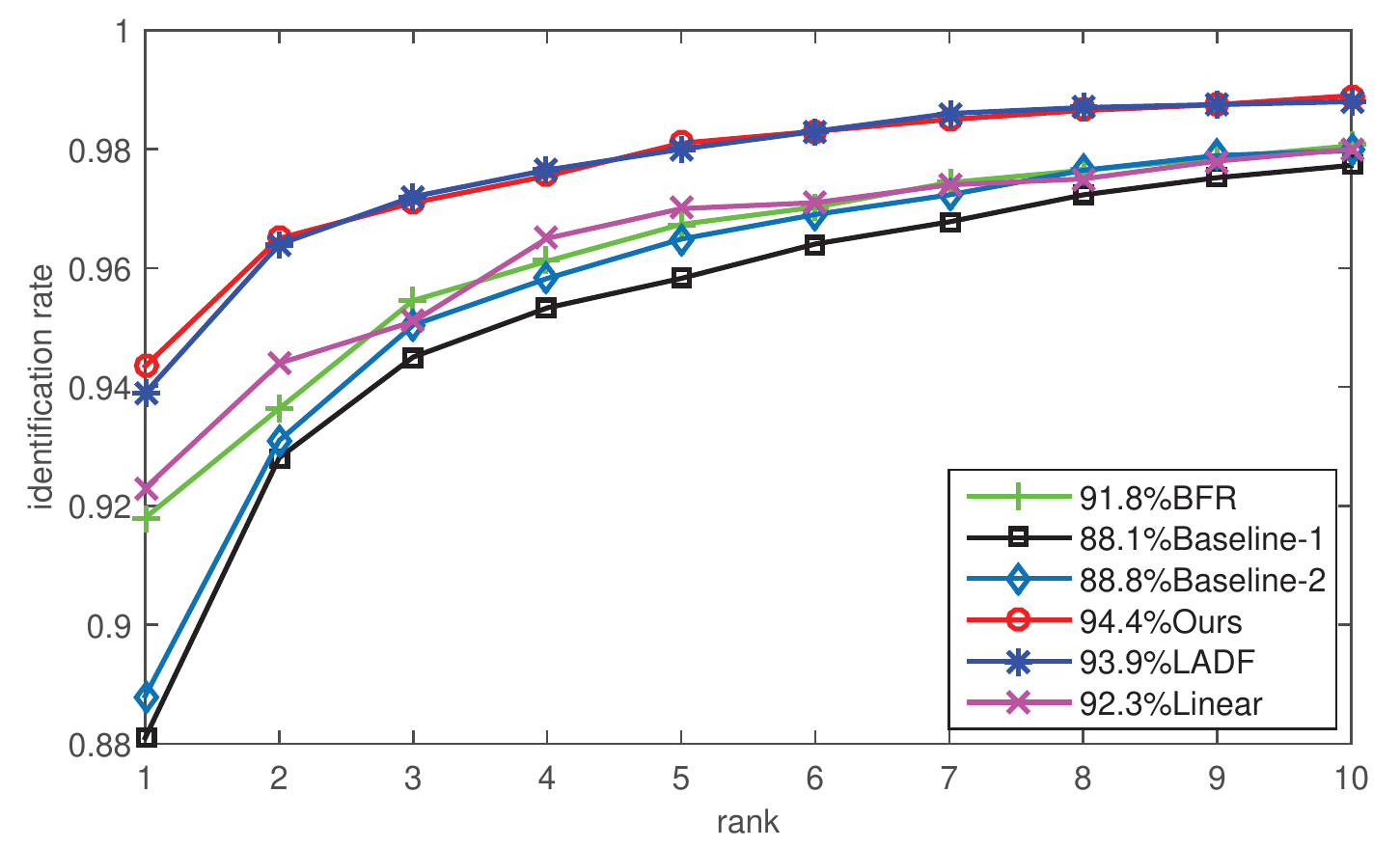}
  \caption{sim.+deep.fea,\\
   MORPH}
  \label{fig:a4}
\end{subfigure}
\begin{subfigure}{0.25\textwidth}
  \centering
  \includegraphics[width=1\linewidth]{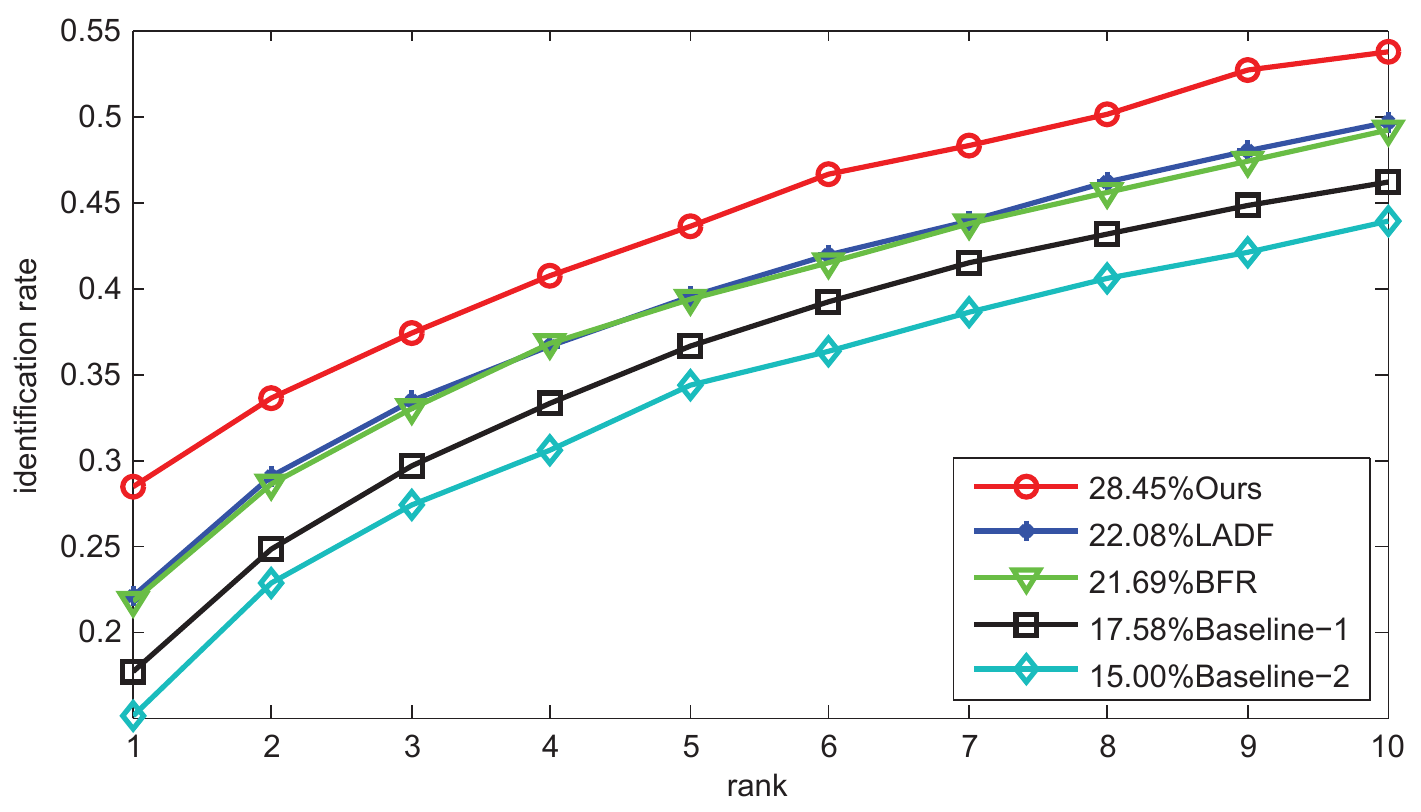}
  \caption{sim.+deep.fea,\\
   COX-V2S}
  \label{fig:a5}
\end{subfigure}%
\begin{subfigure}{0.25\textwidth}
  \centering
  \includegraphics[width=1\linewidth]{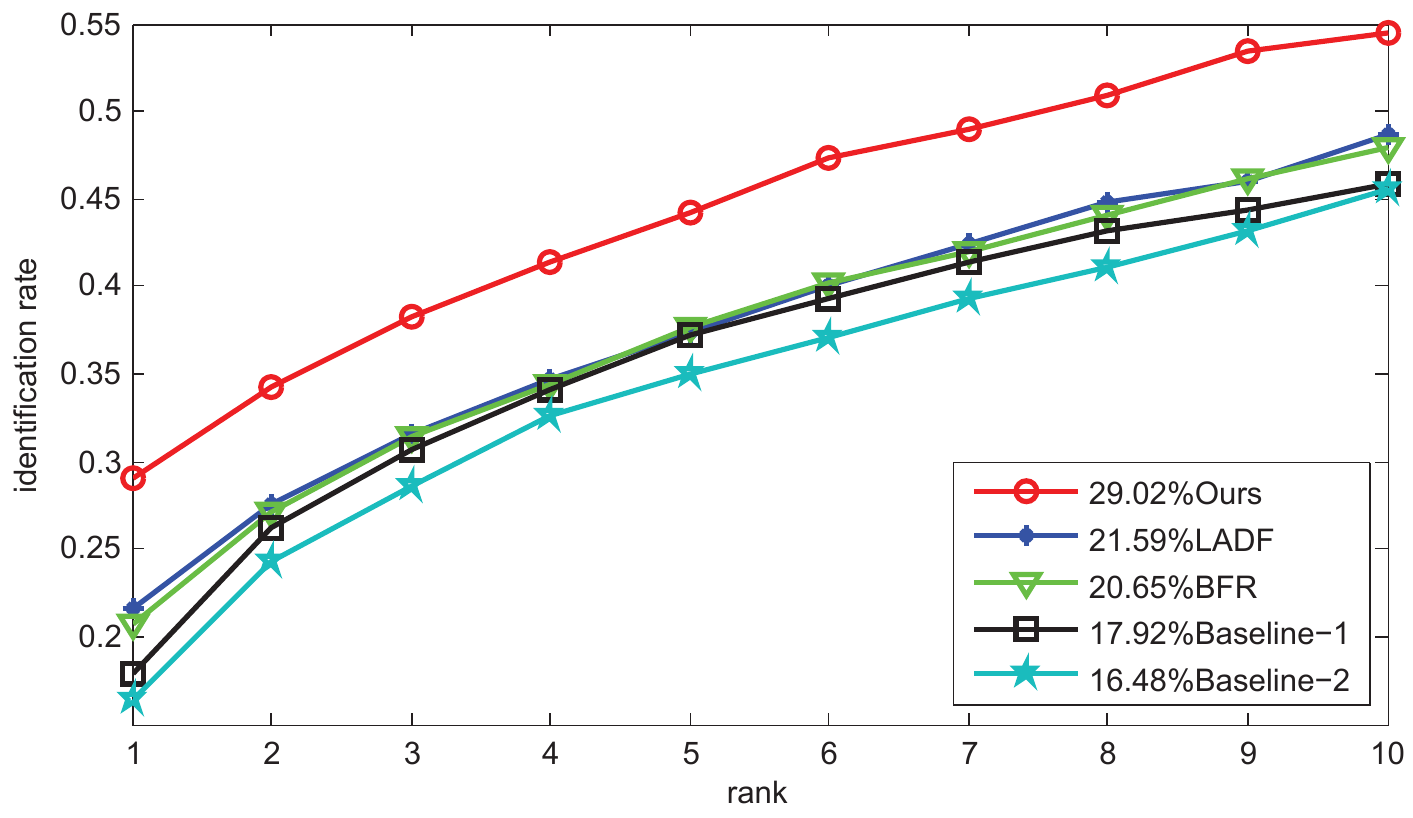}
  \caption{sim.+deep.fea,\\
   COX-S2V}
  \label{fig:a6}
\end{subfigure}
\hfill
\begin{subfigure}{0.23\textwidth}
  \centering
  \includegraphics[width=1\linewidth]{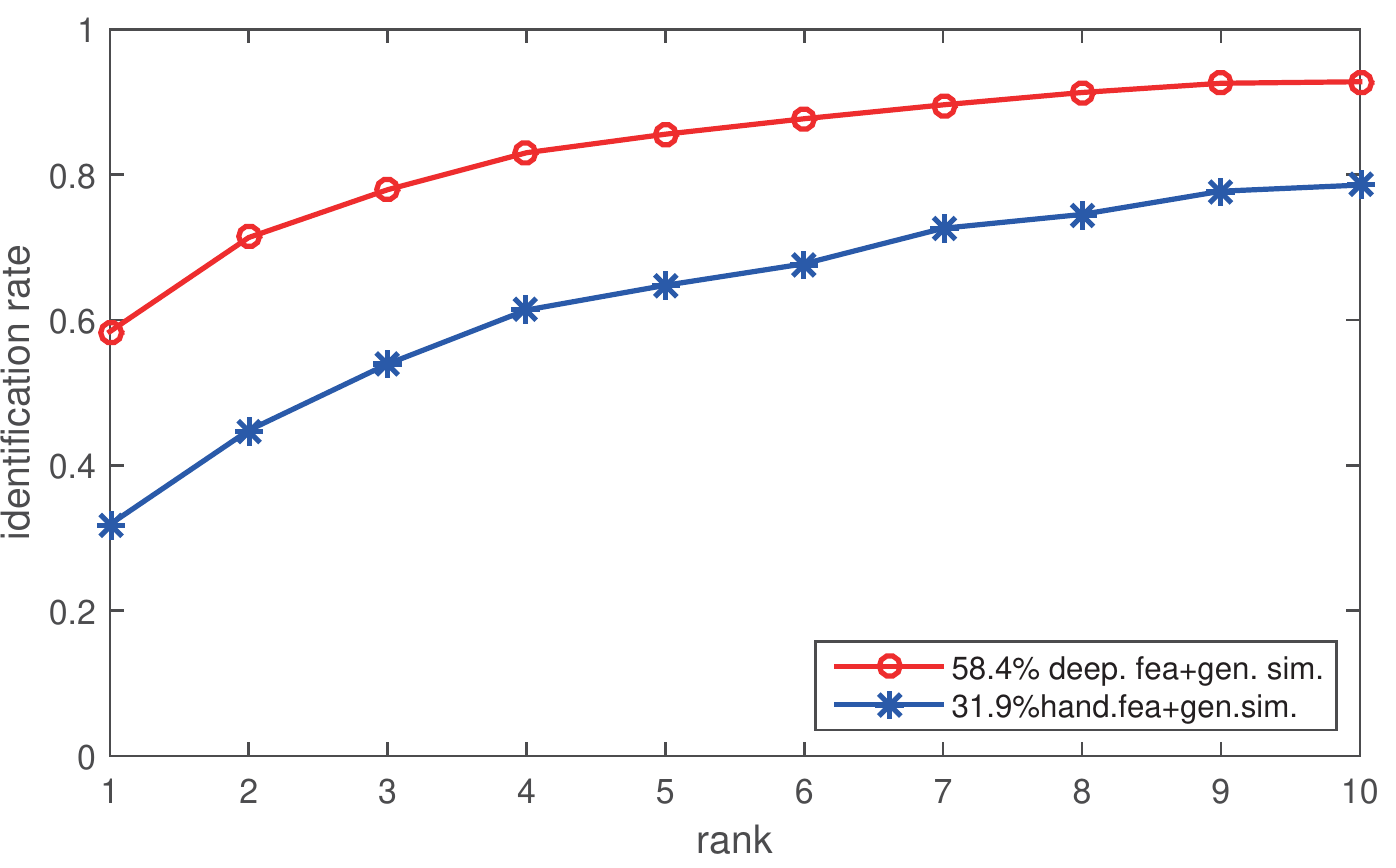}
  \caption{deep/hand fea,\\
   CUHK03}
  \label{fig:a7}
\end{subfigure}%
\begin{subfigure}{0.23\textwidth}
  \centering
  \includegraphics[width=1\linewidth]{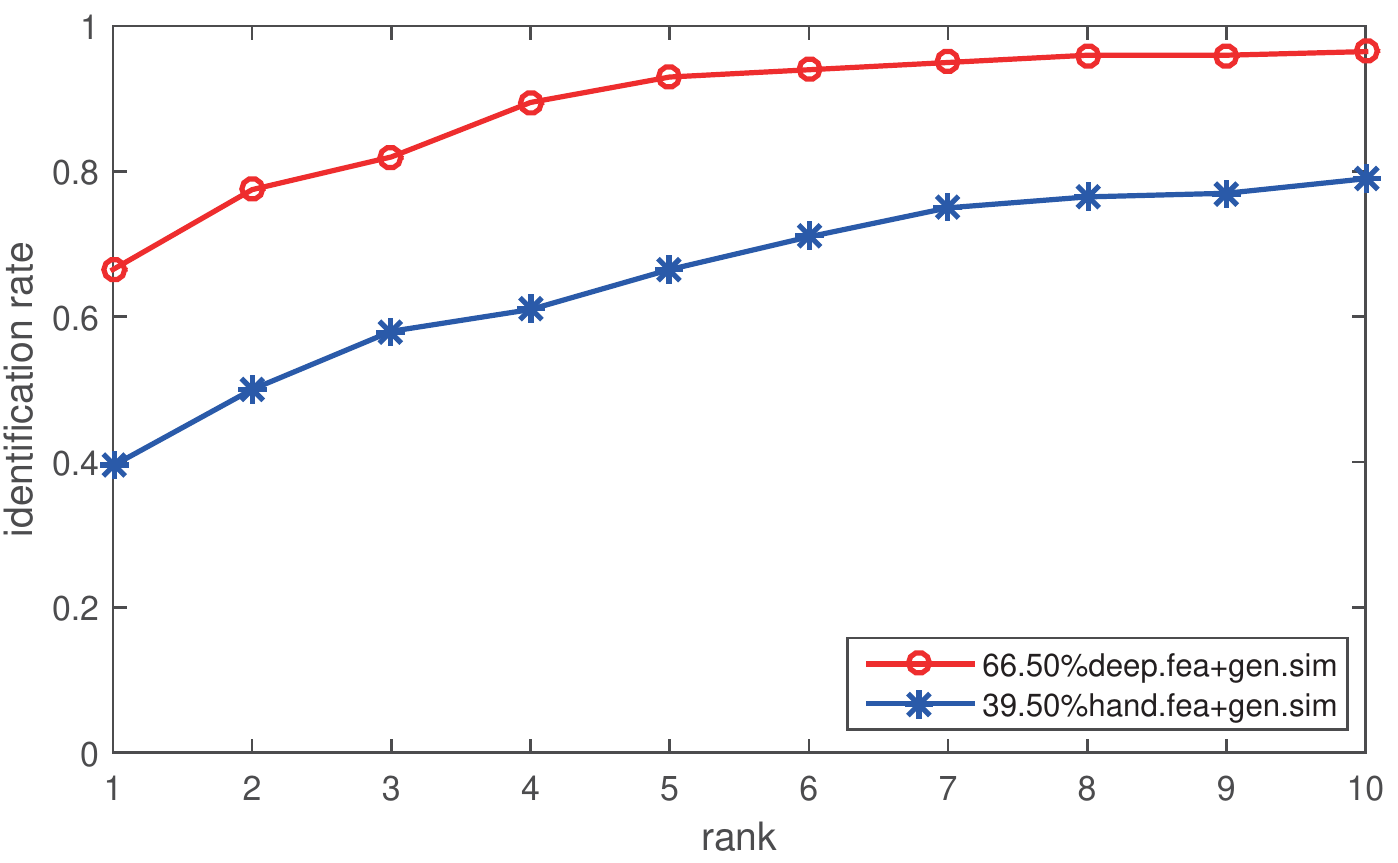}
  \caption{deep/hand fea,\\
   CUHK01}
  \label{fig:a8}
\end{subfigure}
\caption{{Results of the ablation studies demonstrating the effectiveness of each main component of our framework. The CMC curve and  recognition rate are used for evaluation. The results of different similarity models are shown using the handcrafted features (in (a) and (b)) and using the deep features (in (c) - (f) ), respectively. (g) and (h) show the performances with / without the deep feature learning while keeping the same similarity model.}}
\label{fig:comMetricLearning}
\end{figure*}

All of the photos/sketches are resized to $250 \times 200$, and cropped to the size of $230 \times 180$ at the center with a small random perturbation. 1200 pairs of photos and sketches (i.e., including 30 individuals with each having 40 pairs) are constructed for each iteration during the model training. In the testing stage, we use face photos to form the gallery set and treat sketches as the probes.

We employ several existing approaches for comparison: the eigenface transformation based method (ET) \cite{tang2004face}, the multi-scale Markov random field based method (MRF) \cite{wang2009face}, and MRF+ \cite{zhang2010lighting} (i.e., the lighting and pose robust version of \cite{wang2009face}). It is worth mentioning that all of these competing methods need to first synthesize face sketches by photo-sketch transformation, and then measure the similarity between the synthesized sketches and the candidate sketches, while our approach works in an end-to-end way. The quantitative results are reported in Table \ref{tab:12}. Our method achieves 100\% recognition rate on this dataset.

\begin{table}
\caption{Recognition rates on the CUFS dataset for sketch-photo face verification. } \label{tab:12}
\begin{center}
\begin{tabular}{|l|c|}
\hline
Method & Recognition rate\\
\hline\hline
ET \cite{tang2004face} & 71.0\%\\
MRF \cite{wang2009face} & 96.0\%\\
MRF+ \cite{zhang2010lighting} & 99.0\% \\
Ours & \textbf{100.0\%} \\
\hline
\end{tabular}
\end{center}
\end{table}

\subsection{Still-video Face Recognition}

%
\begin{table}
\caption{{Recognition rates on the COX face dataset.}}\label{tab:16}
\begin{center}
\begin{tabular}{|l|c|c|}
\hline
Method & V2S & S2V \\
\hline\hline
PSD \cite{wang2004dual} & 9.90\% & 11.64\% \\
PMD \cite{wang2012manifold} & 6.40\% & 6.10\% \\
PAHD \cite{vincent2001k} & 4.70\% & 6.34\% \\
PCHD \cite{cevikalp2010face} & 7.93\% & 8.89\% \\
PSDML \cite{zhu2013point} & 12.14\% & 7.04\% \\
PSCL-EA  \cite{CoxFace} & \textbf{30.33}\% & 28.39\% \\
Ours & {28.45\%} & \textbf{29.02\%} \\
\hline
\end{tabular}
\end{center}
\end{table}

{Matching person faces across still images and videos is a newly rising task in intelligent visual surveillance. In these applications, the still images (e.g., ID photos) are usually captured under a controlled environment while the faces in surveillance videos are acquired under complex scenarios (e.g., various lighting conditions, occlusions and low resolutions).}

For this task, a large-scale still-video face recognition dataset, namely COX face dataset, has been released recently\footnote{The COX face DB is collected by Institute of Computing Technology Chinese Academy of Sciences, OMRON Social Solutions Co. Ltd, and Xinjiang University.}, which is an extension of the COX-S2V dataset \cite{huang2013benchmarking}. This COX face dataset includes 1,000 subjects and each has one high quality still image and 3 video cliques respectively captured from 3 cameras. {Since these cameras are deployed under similar environments ( e.g., similar results are generated for the three cameras in \cite{CoxFace}, we use the data captured by the first camera in our experiments.}

Following the setting of COX face dataset, we divide the data into a training set (300 subjects) and a testing set (700 subjects), and conduct the experiments with 10 random splits. There are two sub-tasks in the testing: i) matching video frames to still images (V2S) and ii) matching still images to video frames (S2V). {For V2S task we use the video frames as probes and form the gallery set by the still images, and inversely for S2V task.} The split of gallery/probe sets is also consistent with the protocol required by the creator. All of the image are resized to $200 \times 150$, and cropped to the size of $180 \times 130$ with a small random perturbation. 1200 pairs of still images and video frames (i.e., including 20 individuals with each having 60 pairs) are constructed for each iteration during the model training.

Unlike the traditional image-based verification problems, both V2S and S2V are defined as the point-to-set matching problem, i.e., one still image to several video frames (i.e., 10 sampled frames). In the evaluation, we calculate the distance between the still image and each video frame by our model and output the average value over all of the distances. For comparison, we employ several existing point-to-set distance metrics: dual-space linear discriminant analysis (PSD) \cite{wang2004dual}, manifold-manifold distance (PMD) \cite{wang2012manifold}, hyperplane-based distance (PAHD) \cite{vincent2001k}, kernelized convex geometric distance (PCHD) \cite{cevikalp2010face}, and covariance kernel based distance (PSDML) \cite{zhu2013point}. We also compare with the point-to-set correlation learning (PSCL-EA) method \cite{CoxFace}, which specially developed for the COX face dataset. The recognition rates of all competing methods are reported in Table \ref{tab:16}, and our method achieves excellent performances, i.e., the best in S2V and the second best in V2S. {The experiments show that our approach can generally improve performances in the applications to image-to-image, image-to-video, and video-to-image matching problems.}

{\subsection{Ablation Studies}

In order to provide more insights on the performance of our approach, we conduct a number of ablation studies by isolating each main component (e.g., the generalized similarity measure and feature learning). {Besides, we also study the effect of using sample-pair-based and sample-based batch settings in term of convergence efficiency.}

{\bf Generalized Similarity Model.} We design two experiments by using handcrafted features and deep features, respectively, to justify the effectiveness of our generalized similarity measure.

(i) We test our similarity measure using the fixed handcrafted features for person re-identification. The experimental results on CUHK01 and CUHK03 clearly demonstrate the effectiveness of our model against the other similarity models without counting on deep feature learning. Following \cite{zhao2013unsupervised}, we extract the feature representation by using patch-based color histograms and dense SIFT descriptors. This feature representation is fed into a full connection layer for dimensionality reduction to obtain a 400-dimensional vector. We then invoke the similarity sub-network (described in Section \ref{sec:architecture}) to output the measure. On both CUHK01 and CUHK03, we adopt several representative similarity metrics for comparison, i.e., ITML \cite{davis2007information}, LDM \cite{guillaumin2009you}, LMNN \cite{weinberger2005distance}, and RANK \cite{mcfee2010metric}, using the same feature representation.

The quantitative CMC curves and the recognition rates of all these competing models are shown in Fig. \ref{fig:comMetricLearning} (a) and (b) for CUHK03 and CUHK01, respectively, where ``Generalized'' represents our similarity measure. It is observed that our model outperforms the others by large margins, e.g., achieving the rank-1 accuracy of 31.85\% against 13.51\% by LDM on CUHK03. {Most of these competing methods learn Mahalanobis distance metrics. In contrast, our metric model combines Mahalanobis distance with Cosine similarity in a generic form, leading to a more general and effective solution in matching cross-domain data.}

(ii) On the other hand, we incorporate several representative similarity measures into our deep architecture and jointly optimize these measures with the CNN feature learning. Specifically, {we simplify} our network architecture by removing the top layer (i.e., the similarity model), and measure the similarity in either the Euclidean embedding space (as Baseline-1) or in the inner-product space (as Baseline-2). These two variants can be viewed as two degenerations of our similarity measure (i.e., affine Euclidean distance and affine Cosine similarity). To support our discussions in Section \ref{sec:discussion}, we adopt the two distance metric models LADF \cite{li2013learning} and BFR {(i.e., Joint Bayesian)} \cite{chen2012bayesian} into our deep neural networks. Specifically, we replace our similarity model by the LADF model defined in Eqn. (\ref{eq_LADF}) and the BFR model defined in Eqn. (\ref{eq_BFR}), respectively. Moreover, we implement one more variant (denoted as ``Linear'' in this experiment), which applies similarity transformation parameters with separate linear transformations for each data modality. That is, we remove affine transformation while keeping separate linear transformation by setting ${\bf{d}} = {\bf{0}}$, ${\bf{e}} = {\bf{0}}$ and $f = 0$ in Eqn. \ref{eq_general_sim}. Note that the way of incorporating these metric models into the deep architecture is analogously to our metric model. The experiment is conducted on four benchmarks: CUHK03, MORPH, COX-V2S and COX-S2V, and the results are shown in Figure \ref{fig:comMetricLearning} (c), (d), (e), (f), respectively. Our method outperforms the competing methods by large margins on MORPH and COX face dataset. On CUHK03 (i.e., Fig. \ref{fig:comMetricLearning} (c)), our method achieves the best rank-1 identification rate (i.e., $58.39\%$) among all the methods. In particular, the performance drops by $4\%$ when removing the affine transformation on CUHK03.

{It is interesting to discover that most of these competing methods can be treated as special cases of our model. And our generalized similarity model can fully take advantage of convolutional feature learning by developing the specific deep architecture, and can consistently achieve superior performance over other variational models.}


{\bf Deep Feature Learning.} To show the benefit of deep feature learning, we adopt the handcrafted features (i.e., color histograms and SIFT descriptors) on CUHK01 and CHUK03 benchmark. Specifically, we extract this feature representation based on the patches of pedestrian images and then build the similarity measure for person re-identification. The results on {CUHK03 and CHUK01} are reported in Fig. \ref{fig:comMetricLearning} (g) and (h), respectively. We denote the result by using the handcrafted features as ``hand.fea + gen.sim'' and the result by end-to-end deep feature learning as ``deep.fea + gen.sim''. It is obvious that without deep feature representation the performance drops significantly, e.g., from 58.4\% to 31.85\% on CUHK03 and from 66.5\% to 39.5\% on CUHK01. } {These above results clearly demonstrate the effectiveness of utilizing deep CNNs for discriminative feature representation learning.}

{\bf Sample-pair-based vs. sample-based batch setting.} In addition, we conduct an experiment to compare the sample-pair-based and sample-based in term of convergence efficiency, using the CUHK03 dataset. Specifically, for the sample-based batch setting, we select 600 images from 60 people and construct {60,000} pairs in each training iteration. For the sample-pair-based batch setting, 300 pairs are randomly constructed. Note that each person on CUHK03 has 10 images. Thus, 600 images are included in each iteration and the training time per iteration is almost the same for the both settings. Our experiment shows that in the sample-based batch setting, the model achieves rank-1 accuracy of $58.14\%$ after about 175,000 iterations, while in the other setting the rank-1 accuracy is $46.96\%$ after 300,000 iterations. These results validate the effectiveness of the sample-based form in saving the training cost.


\section{Conclusion}
\label{sec:conclusion}

In this work, we have presented a novel generalized similarity model for cross-domain matching of visual data, which generalizes the traditional two-step methods (i.e., projection and distance-based measure). Furthermore, we integrated our model with the feature representation learning by building a deep convolutional architecture. Experiments were performed on several very challenging benchmark dataset of cross-domain matching. The {results show} that our method outperforms other state-of-the-art approaches.

There are several directions along which we intend to extend this work. The first is to extend our approach for larger scale heterogeneous data (e.g., web and user behavior data), thereby exploring new applications (e.g., rich information retrieval). Second, we plan to generalize the pairwise similarity metric into triplet-based learning for more effective model training.

\appendix
\begin{center}
\textbf{Derivation of Equation (\ref{eq_general_sim})}
\end{center}
{


As discussed in Section \ref{sect:intro}, we extend the two linear projections $\mathbf{U}$ and $\mathbf{V}$ into affine transformations and apply them on samples of different domains, ${\bf{x}}$ and ${\bf{y}}$, respectively. That is, we replace $\mathbf{U}\bf{x}$ and $\mathbf{V}{\bf{y}}$ with ${{\bf{L}}_{\bf{A}}}{\bf{x}} + {\bf{a}}$ and ${{\bf{L}}_{\bf{B}}}{\bf{y}} + {\bf{b}}$, respectively. Then, the affine Mahalanobis distance is defined as:

\vspace{-6mm}
\begin{eqnarray}\label{eq_DM}
{D_{\bf{M}}} &=& \left\| {({{\bf{L}}_{\bf{A}}}{\bf{x}} + {\bf{a}}) - ({{\bf{L}}_{\bf{B}}}{\bf{y}} + {\bf{b}})} \right\|_{\bf{2}}^{\bf{2}}\\\nonumber
&=& \begin{bmatrix}\mathbf{x}^T \mbox{ } \mathbf{y}^T \mbox{ } 1\end{bmatrix}
{\mathbb{S}}_{\mathbf{M}}
\begin{bmatrix} \mathbf{x}\\
\mathbf{y}\\
1
\end{bmatrix}.
\end{eqnarray}
where the matrix ${{\bf{\mathbb{S}}}_{\bf{M}}}$ can be further unfolded as:

\vspace{-3mm}
\begin{equation}\label{eq_mmatrix}
{{\bf{\mathbb{S}}}_{\mathbf{M}}} = \left[ {\begin{array}{*{20}{c}}
{{\bf{L}}_{\bf{A}}^T{{\bf{L}}_{\bf{A}}}}&{ - {\bf{L}}_{\bf{A}}^T{{\bf{L}}_{\bf{B}}}}&{{\bf{L}}_{\bf{A}}^T({\bf{a}} - {\bf{b}})}\\
{ - {\bf{L}}_{\bf{B}}^T{{\bf{L}}_{\bf{A}}}}&{{\bf{L}}_{\bf{B}}^T{{\bf{L}}_{\bf{B}}}}&{{\bf{L}}_{\bf{B}}^T({\bf{b}} - {\bf{a}})}\\
{({\bf{a}}^T - {\bf{b}}^T){{\bf{L}}_{\bf{A}}}}&{({\bf{b}}^T - {\bf{a}}^T){{\bf{L}}_{\bf{B}}}}&{\left\| {{{\bf{a}}} - {{\bf{b}}}} \right\|_{\bf{2}}^{\bf{2}}}
\end{array}} \right].
\end{equation}

Furthermore, the affine Cosine similarity is defined as the inner product in the space of affine transformations:

\vspace{-4mm}
\begin{eqnarray}\label{eq_DI}
{S_{\bf{I}}} &=&  {({{{\bf{\mathord{\buildrel{\lower3pt\hbox{$\scriptscriptstyle\frown$}}
\over L} }}}_{\bf{A}}}{\bf{x}} + {{{\bf{\mathord{\buildrel{\lower3pt\hbox{$\scriptscriptstyle\frown$}}
\over {\bf{a}}} }}}})^T ({{{\bf{\mathord{\buildrel{\lower3pt\hbox{$\scriptscriptstyle\frown$}}
\over L} }}}_{\bf{B}}}{\bf{y}} + {{{\bf{\mathord{\buildrel{\lower3pt\hbox{$\scriptscriptstyle\frown$}}
\over {\bf{b}}} }}}})} \\\nonumber
&=& \begin{bmatrix}\mathbf{x}^T \mbox{ } \mathbf{y}^T \mbox{ } 1\end{bmatrix}
{\mathbb{S}}_{\mathbf{I}}
\begin{bmatrix} \mathbf{x}\\
\mathbf{y}\\
1
\end{bmatrix}.
\end{eqnarray}
The corresponding matrix ${{\bf{\mathbb{S}}}_{\bf{I}}} $ is,

\vspace{-4mm}
\begin{equation}\label{eq_imatrix}
{{\bf{\mathbb{S}}}_{\bf{I}}} = \left[ {\begin{array}{*{20}{c}}
{\bf{0}}&{\frac{{{\bf{\mathord{\buildrel{\lower3pt\hbox{$\scriptscriptstyle\frown$}}
\over L} }}_{\bf{A}}^T{{{\bf{\mathord{\buildrel{\lower3pt\hbox{$\scriptscriptstyle\frown$}}
\over L} }}}_{\bf{B}}}}}{2}}&{\frac{{{\bf{\mathord{\buildrel{\lower3pt\hbox{$\scriptscriptstyle\frown$}}
\over L} }}_{\bf{A}}^T{{{\bf{\mathord{\buildrel{\lower3pt\hbox{$\scriptscriptstyle\frown$}}
\over {\bf{b}}} }}}}}}{2}}\\
{\frac{{{\bf{\mathord{\buildrel{\lower3pt\hbox{$\scriptscriptstyle\frown$}}
\over L} }}_{\bf{B}}^T{{{\bf{\mathord{\buildrel{\lower3pt\hbox{$\scriptscriptstyle\frown$}}
\over L} }}}_{\bf{A}}}}}{2}}&{\bf{0}}&{\frac{{{\bf{\mathord{\buildrel{\lower3pt\hbox{$\scriptscriptstyle\frown$}}
\over L} }}_{\bf{B}}^T{{{\bf{\mathord{\buildrel{\lower3pt\hbox{$\scriptscriptstyle\frown$}}
\over {\bf{a}}} }}}}}}{2}}\\
{\frac{{{\bf{\mathord{\buildrel{\lower3pt\hbox{$\scriptscriptstyle\frown$}}
\over {\bf{b}}} }}^T{{{\bf{\mathord{\buildrel{\lower3pt\hbox{$\scriptscriptstyle\frown$}}
\over L} }}}_{\bf{A}}}}}{2}}&{\frac{{{\bf{\mathord{\buildrel{\lower3pt\hbox{$\scriptscriptstyle\frown$}}
\over {\bf{a}}} }}^T{{{\bf{\mathord{\buildrel{\lower3pt\hbox{$\scriptscriptstyle\frown$}}
\over L} }}}_{\bf{B}}}}}{2}}&{{\bf{\mathord{\buildrel{\lower3pt\hbox{$\scriptscriptstyle\frown$}}
\over {\bf{a}}} }}^T{{{\bf{\mathord{\buildrel{\lower3pt\hbox{$\scriptscriptstyle\frown$}}
\over {\bf{b}}} }}}}}
\end{array}} \right]{\kern 1pt} {\kern 1pt} {\kern 1pt} ,
\end{equation}

We propose to fuse ${D_{\bf{M}}}$ and ${S_{\bf{I}}}$ by a weighted aggregation as follows:

\vspace{-5mm}
\begin{eqnarray}\label{eq_combinedsimilarity}
{S} &=& \mu {D_{\bf{M}}} - \lambda {S_{\bf{I}}} \\\nonumber
&=& \begin{bmatrix}\mathbf{x}^T \mbox{ } \mathbf{y}^T \mbox{ } 1\end{bmatrix}
{\bf{\mathbb{S}}}
\begin{bmatrix} \mathbf{x}\\
\mathbf{y}\\
1
\end{bmatrix}.
\end{eqnarray}
{Note that ${D_{\bf{M}}}$ is an affine distance (i.e., nonsimilarity) measure while ${S_{\bf{I}}}$ is an affine similarity measure. Analogous to \cite{cao2013similarity}, we adopt $\mu {D_{\bf{M}}} - \lambda {S_{\bf{I}}}$ ($\mu, \lambda \geq 0$) to combine ${D_{\bf{M}}}$ and ${S_{\bf{I}}}$.} The parameters $\mu$ , $\lambda$, ${D_{\bf{M}}}$ and ${S_{\bf{I}}}$ are automatically learned through our learning algorithm. Then, the matrix ${\bf{\mathbb{S}}}$ can be obtained by fusing ${{\bf{\mathbb{S}}}_{\bf{M}}}$ and ${{\bf{\mathbb{S}}}_{\bf{I}}}$:

\vspace{-2mm}
\begin{equation}\label{eq_combinedmatrix}
{\mathbb{S}} = \left[ {\begin{array}{*{20}{c}}
{{\mathbf{A}}}&{{\mathbf{C}}}&{{\mathbf{d}}}\\
{{{\mathbf{C}}^T}}&{{\mathbf{B}}}&{{\mathbf{e}}}\\
{{{\mathbf{d}}^T}}&{{{\mathbf{e}}^T}}&{f}
\end{array}} \right],
\end{equation}
where

\vspace{-4mm}
\begin{equation}\label{eq_abcdef}
\begin{array}{*{20}{c}}
{{\bf{A}} = \mu {\bf{L}}_{\bf{A}}^T{{\bf{L}}_{\bf{A}}}}\\
{}\\
{{\bf{B}} = \mu {\bf{L}}_{\bf{B}}^T{{\bf{L}}_{\bf{B}}}}\\
{}\\
{{\bf{C}} = {\bf{ - }}\mu {\bf{L}}_{\bf{A}}^T{{\bf{L}}_{\bf{B}}} - \lambda \frac{{{\bf{\mathord{\buildrel{\lower3pt\hbox{$\scriptscriptstyle\frown$}}
\over L} }}_{\bf{A}}^T{{{\bf{\mathord{\buildrel{\lower3pt\hbox{$\scriptscriptstyle\frown$}}
\over L} }}}_{\bf{B}}}}}{2}}\\
{}\\
{{\bf{d}} = \mu {{\bf{L}}_{\bf{A}}^T}({{\bf{a}}}{\bf{ - }}{{\bf{b}}}) - \lambda \frac{{{\bf{\mathord{\buildrel{\lower3pt\hbox{$\scriptscriptstyle\frown$}}
\over L} }}_{\bf{A}}^T{{{\bf{\mathord{\buildrel{\lower3pt\hbox{$\scriptscriptstyle\frown$}}
\over {\bf{b}}} }}}}}}{2}}\\
{}\\
{{\bf{e}} = \mu {{\bf{L}}_{\bf{B}}^T}({{\bf{b}}}{\bf{ - }}{{\bf{a}}}) - \lambda \frac{{{\bf{\mathord{\buildrel{\lower3pt\hbox{$\scriptscriptstyle\frown$}}
\over L} }}_{\bf{B}}^T{{{\bf{\mathord{\buildrel{\lower3pt\hbox{$\scriptscriptstyle\frown$}}
\over {\bf{a}}} }}}}}}{2}}\\
{}\\
{f = \mu \left\| {{{\bf{a}}}{\bf{ - }}{{\bf{b}}}} \right\|_{\bf{2}}^{\bf{2}} - \lambda {\bf{\mathord{\buildrel{\lower3pt\hbox{$\scriptscriptstyle\frown$}}
\over {\bf{a}}} }}^T{{{\bf{\mathord{\buildrel{\lower3pt\hbox{$\scriptscriptstyle\frown$}}
\over {\bf{b}}} }}}}}
\end{array}.
\end{equation}
In the above equations, we use $6$ matrix (vector) variables, i.e., ${\bf{A}}$, ${\bf{B}}$, ${\bf{C}}$, ${\bf{d}}$, ${\bf{e}}$ and $f$, to represent the parameters of the generalized similarity model in a generic form. On one hand, given $\mu$, $\lambda$, ${{\bf{\mathbb{S}}}_{\bf{M}}}$ and ${{\bf{\mathbb{S}}}_{\bf{I}}}$, these matrix variables can be directly determined using Eqn. (\ref{eq_abcdef}). On the other hand, if we impose the positive semi-definite constraint on ${\bf{A}}$ and ${\bf{B}}$, it can be proved that once ${\bf{A}}$, ${\bf{B}}$, ${\bf{C}}$, ${\bf{d}}$, ${\bf{e}}$ and $f$ are determined there exist at least one solution of $\mu$, $\lambda$, ${{\bf{\mathbb{S}}}_{\bf{M}}}$ and ${{\bf{\mathbb{S}}}_{\bf{I}}}$, respectively, that is, $\mathbb{S}$ is guaranteed to be decomposed into the weighted Mahalanobis distance and Cosine similarity. Therefore, the generalized similarity measure can be learned by optimizing ${\bf{A}}$, ${\bf{B}}$, ${\bf{C}}$, ${\bf{d}}$, ${\bf{e}}$ and $f$ under the positive semi-definite constraint on ${\bf{A}}$ and ${\bf{B}}$. In addition, ${\bf{C}}$ is not required to satisfy the positive semidefinite condition and it may not be a square matrix when the dimensions of ${\bf{x}}$ and ${\bf{y}}$ are unequal.

}

\ifCLASSOPTIONcaptionsoff
  \newpage
\fi




\section*{Acknowledgment}
This work was supported in part by Guangdong Natural Science Foundation under Grant S2013050014548 and 2014A030313201, in part by Program of Guangzhou Zhujiang Star of Science and Technology under Grant 2013J2200067, and in part by the Fundamental Research Funds for the Central Universities. This work was also supported by Special Program for Applied Research on Super Computation of the NSFC-Guangdong Joint Fund (the second phase).

\bibliographystyle{IEEEtran}
\bibliography{mybibfile}




\begin{IEEEbiography}[{\includegraphics[width=1in,height=1.25in,clip,keepaspectratio]{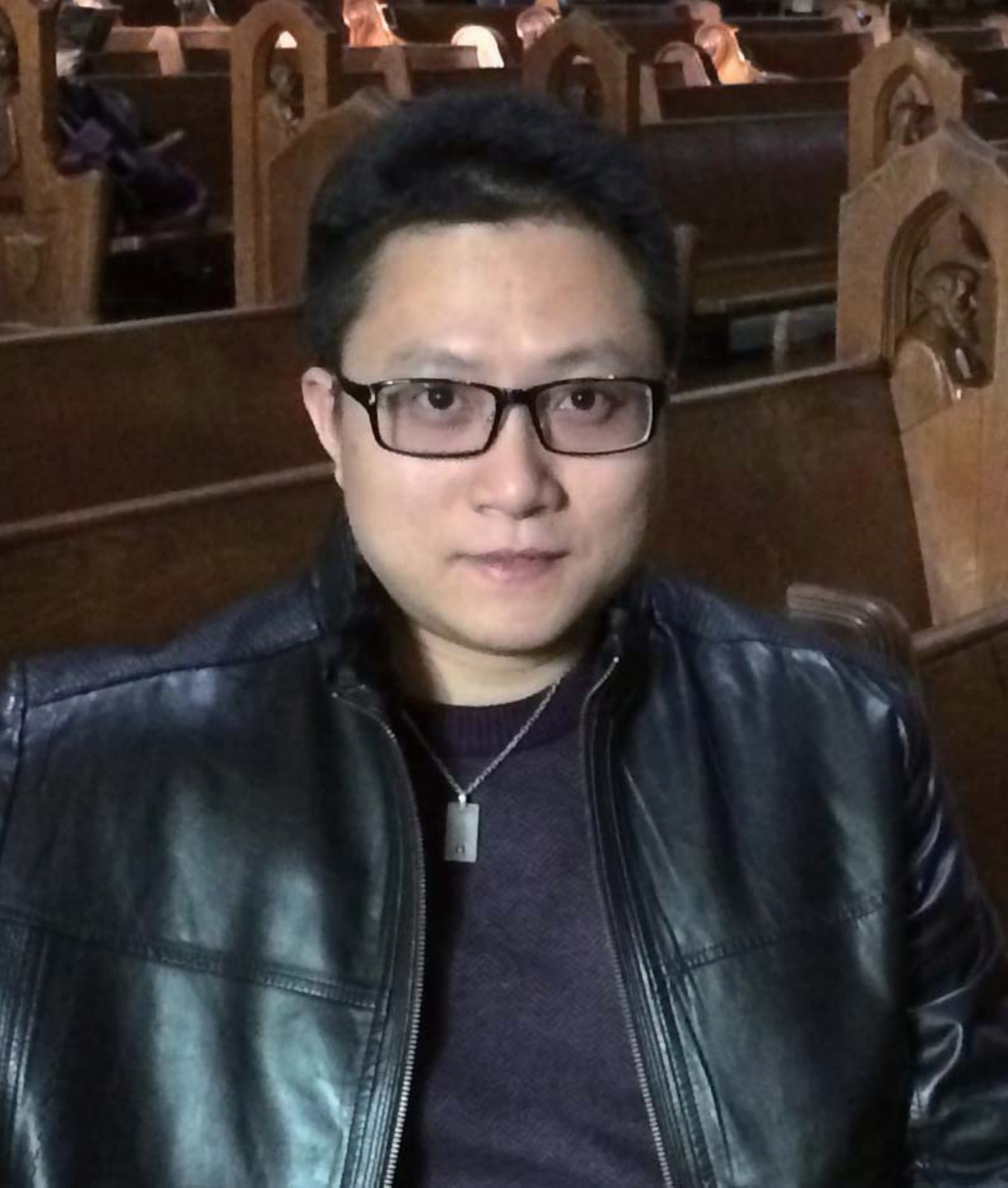}}]{Liang Lin} is a Professor with the School of computer science, Sun Yat-Sen University (SYSU), China. He received the B.S. and Ph.D. degrees from the Beijing Institute of Technology (BIT), Beijing, China, in 1999 and 2008, respectively. From 2008 to 2010, he was a Post-Doctoral Research Fellow with the Department of Statistics, University of California, Los Angeles UCLA. He worked as a Visiting Scholar with the Department of Computing, Hong Kong Polytechnic University, Hong Kong and with the Department of Electronic Engineering at the Chinese University of Hong Kong. His research focuses on new models, algorithms and systems for intelligent processing and understanding of visual data such as images and videos. He has published more than 100 papers in top tier academic journals and conferences. He currently serves as an associate editor of IEEE Tran. Human-Machine Systems. He received the Best Paper Runners-Up Award in ACM NPAR 2010, Google Faculty Award in 2012, Best Student Paper Award in IEEE ICME 2014, and Hong Kong Scholars Award in 2014.
\end{IEEEbiography}

\begin{IEEEbiography}[{\includegraphics[width=1in,height=1.25in,clip,keepaspectratio]{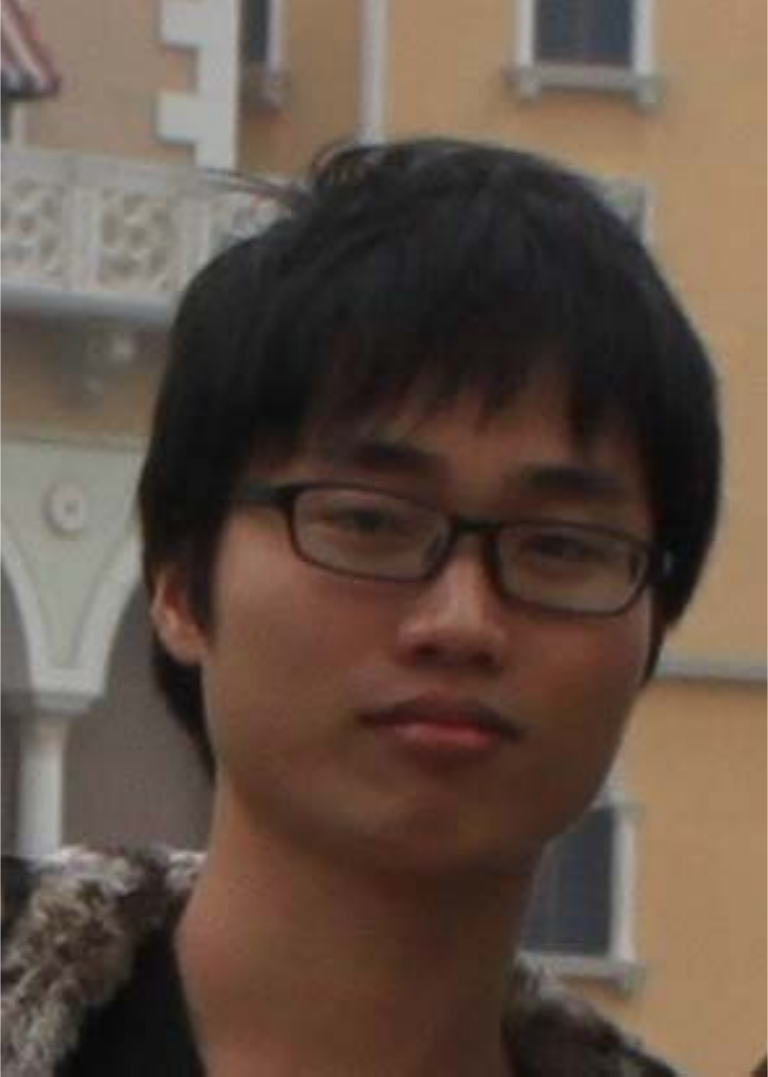}}]{Guangrun Wang} received the B.E. degree from the School of Information Science and Technology, Sun Yat-sen University, Guangzhou, China, in 2013. He is currently pursuing the M.E. degree in the School of Data and Computer Science, Sun Yat-sen University. His research interests include computer vision and machine learning.
\end{IEEEbiography}

\begin{IEEEbiography}[{\includegraphics[width=1in,height=1.25in,clip,keepaspectratio]{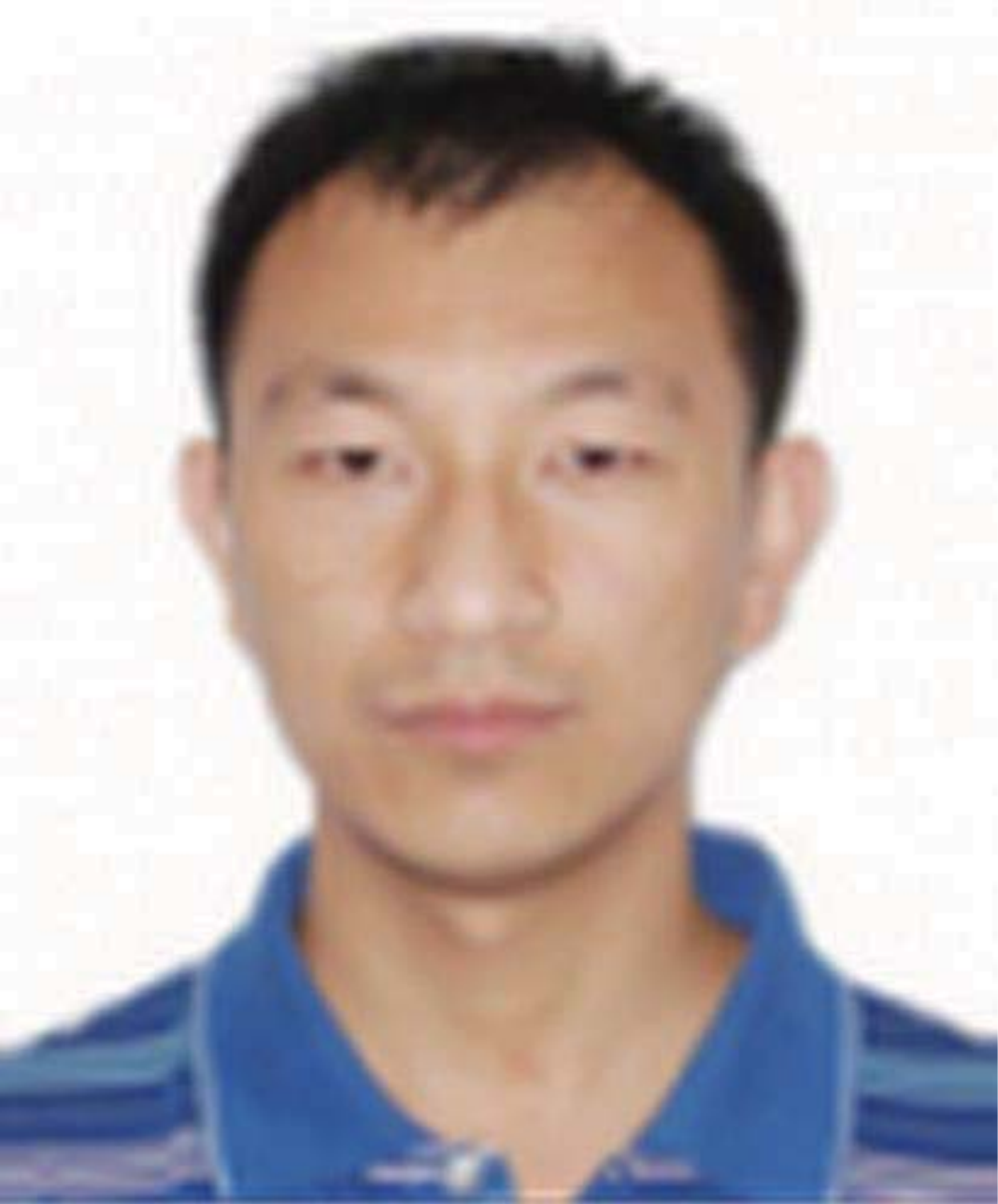}}]{Wangmeng Zuo}
(M'09, SM'14) received the Ph.D. degree in computer application technology from the Harbin Institute of Technology, Harbin, China, in 2007. In 2004, from 2005 to 2006, and from 2007 to 2008, he was a Research Assistant with the Department of Computing, Hong Kong Polytechnic University, Hong Kong. From 2009 to 2010, he was a Visiting Professor at Microsoft Research Asia. He is
currently an Professor with the School of Computer Science and Technology, Harbin Institute of Technology. His current research interests include image modeling and low-level vision, discriminative learning, and biometrics. He has authored about 50 papers in those areas. He is an Associate Editor of the IET Biometrics.
\end{IEEEbiography}

\begin{IEEEbiography}[{\includegraphics[width=1in,height=1.25in,clip,keepaspectratio]{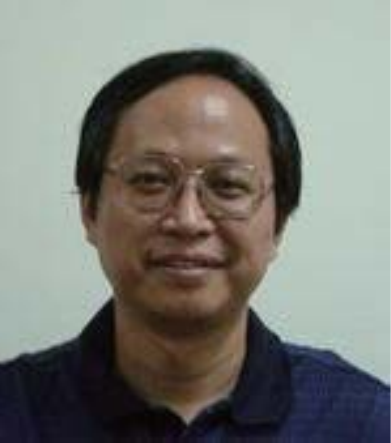}}]{Xiangchu Feng} received the B.E. degree in computational mathematics from Xi'an Jiaotong University and the M.S. and Ph.D. degree in applied mathematics from Xidian University, Xi'an, China in 1984, 1989 and 1999, respectively. Currently, he is a Professor in the Department of Information and Computational Science, School of Math. and Statistics, Xidian University, Xi'an, China. His current research interests include advanced numerical analysis, image restoration and enhancement based on PDEs and sparse approximation.
\end{IEEEbiography}

\begin{IEEEbiography}[{\includegraphics[width=1in,height=1.25in,clip,keepaspectratio]{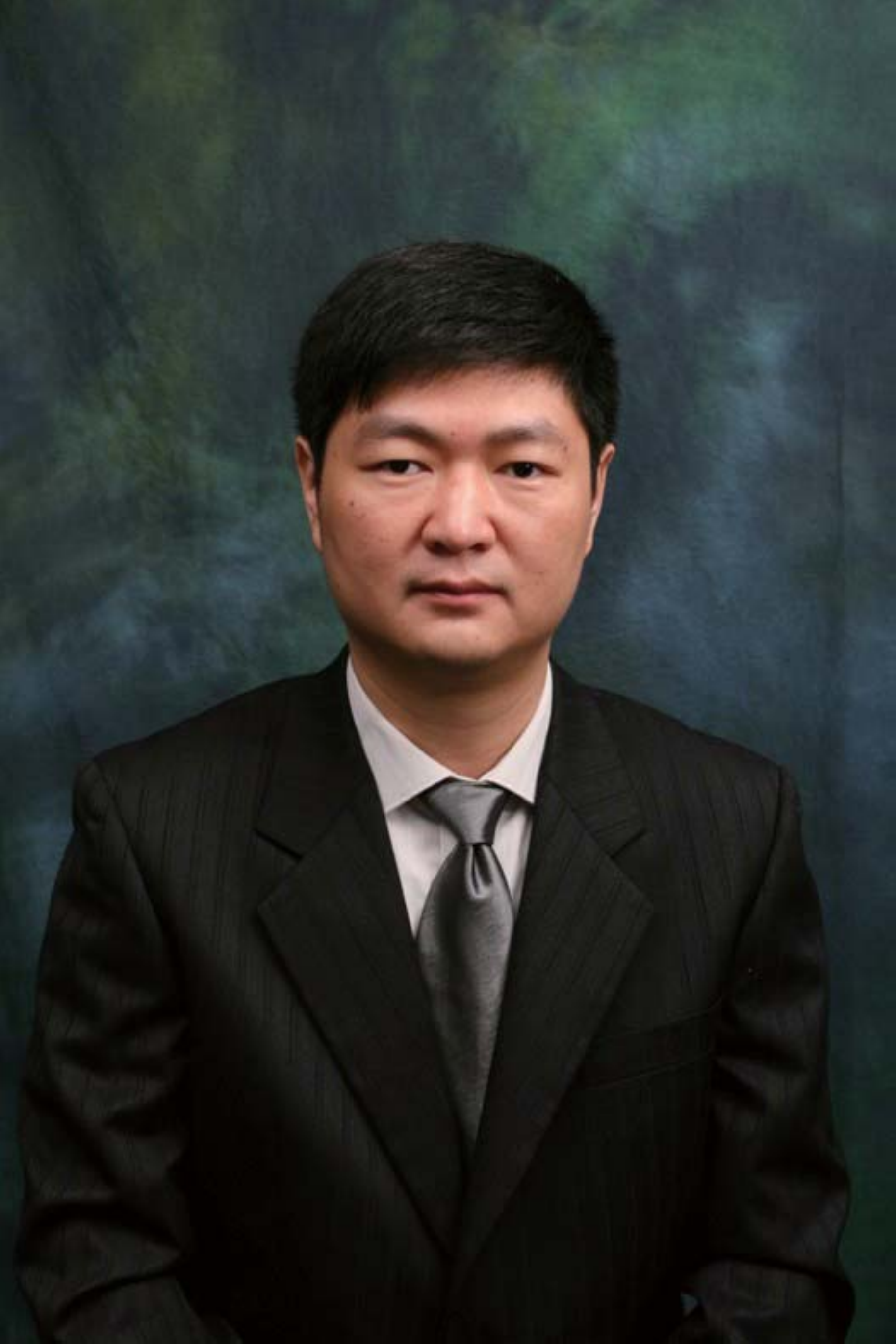}}]{Lei Zhang}
(M'04, SM'14) received the B.Sc. degree in 1995 from Shenyang Institute of Aeronautical Engineering, Shenyang, P.R. China, the M.Sc. and Ph.D degrees in Control Theory and Engineering from Northwestern Polytechnical University, Xi¡¯an, P.R. China, respectively in 1998 and 2001. From 2001 to 2002, he was a research associate in the Dept. of Computing, The Hong Kong Polytechnic University. From Jan. 2003 to Jan. 2006 he worked as a Postdoctoral Fellow in the Dept. of Electrical and Computer Engineering, McMaster University, Canada. In 2006, he joined the Dept. of Computing, The Hong Kong Polytechnic University, as an Assistant Professor. Since July 2015, he has been a Full Professor in the same department. His research interests include Computer Vision, Pattern Recognition, Image and Video Processing, and Biometrics, etc. Dr. Zhang has published more than 200 papers in those areas. By 2015, his publications have been cited more than 14,000 times in literature. Dr. Zhang is currently an Associate Editor of IEEE Trans. on Image Processing, IEEE Trans. on CSVT and Image and Vision Computing. He was awarded the 2012-13 Faculty Award in Research and Scholarly Activities. More information can be found in his homepage http://www4.comp.polyu.edu.hk/~cslzhang/.
\end{IEEEbiography}

\end{document}